\documentclass[a4paper,fleqn]{cas-dc}

\usepackage[numbers]{natbib}
\usepackage{algorithm}
\usepackage[noend]{algpseudocode}
\usepackage{xcolor}
\usepackage{setspace}
\usepackage{siunitx}

\usepackage{amsmath}
\usepackage{amssymb}

\usepackage[export]{adjustbox}
\usepackage[margins]{trackchanges}
\addeditor{AH}

\usepackage{color}

\def\tsc#1{\csdef{#1}{\textsc{\lowercase{#1}}\xspace}}
\tsc{WGM}
\tsc{QE}
\tsc{EP}
\tsc{PMS}
\tsc{BEC}
\tsc{DE}

\begin{document}

\let\WriteBookmarks\relax
\def\floatpagepagefraction{1}
\def\textpagefraction{.001}
\shorttitle{Intelligent Cooperative Robotics in Additive Manufacturing}
\shortauthors{Rescsanski et~al.}

\title [mode = title]{Towards Intelligent Cooperative Robotics in Additive Manufacturing: Past, Present and Future}

\author[1]{Sean Rescsanski}

\author[2]{Rainer Hebert}

\author[3]{Azadeh Haghighi}

\author[1]{Jiong Tang}

\author[1]{Farhad Imani}[
  orcid = 0000-0003-0837-9273,
  ]
\cormark[1]

\ead{farhad.imani@uconn.edu}

\address[1]{School of Mechanical, Aerospace, and Manufacturing Engineering, University of Connecticut, Storrs, CT 06269, USA.}

\address[2]{Department of Materials Science and Engineering, University of Connecticut, Storrs, CT 06269, USA.}

\address[3]{Department of Mechanical and Industrial Engineering, University of Illinois Chicago, Chicago, IL 60608, USA.}

\cortext[cor1]{Corresponding author}

\begin{abstract}
Additive manufacturing (AM) technologies have undergone significant advancements through the integration of cooperative robotics additive manufacturing (C-RAM) platforms. By deploying AM processes on the end effectors of multiple robotic arms, not only are traditional constraints such as limited build volumes circumvented, but systems also achieve accelerated fabrication speeds, cooperative sensing capabilities, and in-situ multi-material deposition.
Despite advancements, challenges remain, particularly regarding defect generation including voids, cracks, and residual stress. Various factors contribute to these issues, including toolpath planning (i.e., slicing strategies), part decomposition for cooperative printing, and motion planning (i.e., path and trajectory planning).
This review first examines the critical aspects of system control for C-RAM systems comprised of slicing and motion planning. The methods for the mitigation of defects through the adjustment of these aspects and the process parameters of AM methods are then described in the context of how they modify the AM process: pre-process, inter-layer (i.e., during layer pauses), and mid-layer (i.e., during material deposition). The application of advanced sensing technologies, including high-resolution cameras, laser scanners, and thermal imaging, to facilitate the capture of micro, meso, and macro-scale defects is explored. The role of digital twins is analyzed, emphasizing their capability to simulate and predict manufacturing outcomes, enabling preemptive adjustments to prevent defects. Finally, the outlook and future opportunities for developing next-generation C-RAM systems are outlined. 

\end{abstract}


    \maketitle


\section{Introduction}


Additive manufacturing (AM) processes utilize layer-by-layer material deposition to fabricate parts from feedstock as opposed to conventional manufacturing which utilize subtractive or forming methods from stock material \cite{gibson2021additive}. First developed in the 1980s \cite{zhai2014additive}, AM has grown to seven recognized categories \cite{ASTM52900-2021}, specifically material extrusion (ME), material jetting, binder jetting, directed energy deposition (DED), sheet lamination, and photo-polymerization \cite{grandviewresearch2024additive}.
These processes utilize a wide range of feedstock materials such as polymers, metals, composites, and ceramics \cite{bhatiaAdditiveManufacturingMaterials2023}. AM's capability to fabricate complex geometries, reduce material waste and decrease production turnaround has driven use in various industrial areas including aerospace \cite{blakey2021metal}, defense \cite{badiru2017additive}, construction \cite{craveiro2019additive, khosravani2022large}, and medical \cite{salmi2021additive} and have led to significant market growth, with expectations to reach over \$70 billion globally by 2030 \cite{grandviewresearch2024additive}.

The majority of AM processes employ gantry-style platforms to afford the control necessary for part fabrication while remaining inexpensive \cite{alhijailyTeamsRobotsAdditive2023, siciliano2008robotica}. 
Several AM processes including powder bed fusion and VAT photopolymerization require gantry-style platforms as a result of process characteristics (e.g., material handling) which constrains system design \cite{singh2021powder, zhang2021recent}.
While effective, the nature of the "in-bounds" build volume (i.e. static, exclusive build regions) of gantries results in a limited maximum part size unsuitable for large-scale part fabrication \cite{vafadar2021advances}. In instances where parts exceed the maximum build volume, they must be decomposed into segments which greatly reduces mechanical strength. 
Furthermore, large-scale gantry systems are prohibitively expensive and have subsequently seen little integration for large-scale AM. 

Recent research has turned to the development of AM systems that leverage robotic arms to facilitate the motion of toolheads, commonly referred to as robotic additive manufacturing (RAM) \cite{zhang2015robotic}, to provide enhanced capabilities and alleviate drawbacks found in conventional gantry-style systems \cite{bhattExpandingCapabilitiesAdditive2020}.
The high degree of freedoms (DoF) of RAM systems affords both multi-plane and non-planar slicing methods which can improve mechanical strength and reduce support material usage \cite{nayyeriPlanarNonplanarSlicing2022, tangReviewMultiaxisAdditive2024}.
The large build area and out-of-bounds build volumes of RAM systems enable the fabrication of large-scale parts with both single and multi-arm configurations that utilize overlapping build volumes \cite{vicenteLargeformatAdditiveManufacturing2023}. RAM systems have been explored in a variety of AM processes including ME \cite{badarinathIntegrationEvaluationRobotic2021}, DED \cite{hassenScalingMetalAdditive2020} and stereolithography \cite{stevensConformalRoboticStereolithography2016}. However, wire-fed DED processes such as wire arc additive manufacturing (WAAM) suffer from defects such as residual stress, porosity, and void generation \cite{svetlizky2021directed}.
Powder-fed DED processes such as laser-DED (L-DED) similarly suffer from porosity and voids in addition to powder-specific defects including keyholing \cite{liu2021review}.
ME processes such as fused filament fabrication (FFF)/fused deposition modeling (FDM) suffer from defects including internal porosity, often known as voids, which form in between deposition tracks within layers \cite{sun_review_2023}. The presence of these defects can require the re-manufacturing of RAM components leading to massive waste of time and costs especially for large-scale parts.



\begin{figure*}[h!]
    \centering
    \begin{tabular}{c c}
        \begin{minipage}[t]{0.35\textwidth}
            \centering
            \includegraphics[scale=0.25]{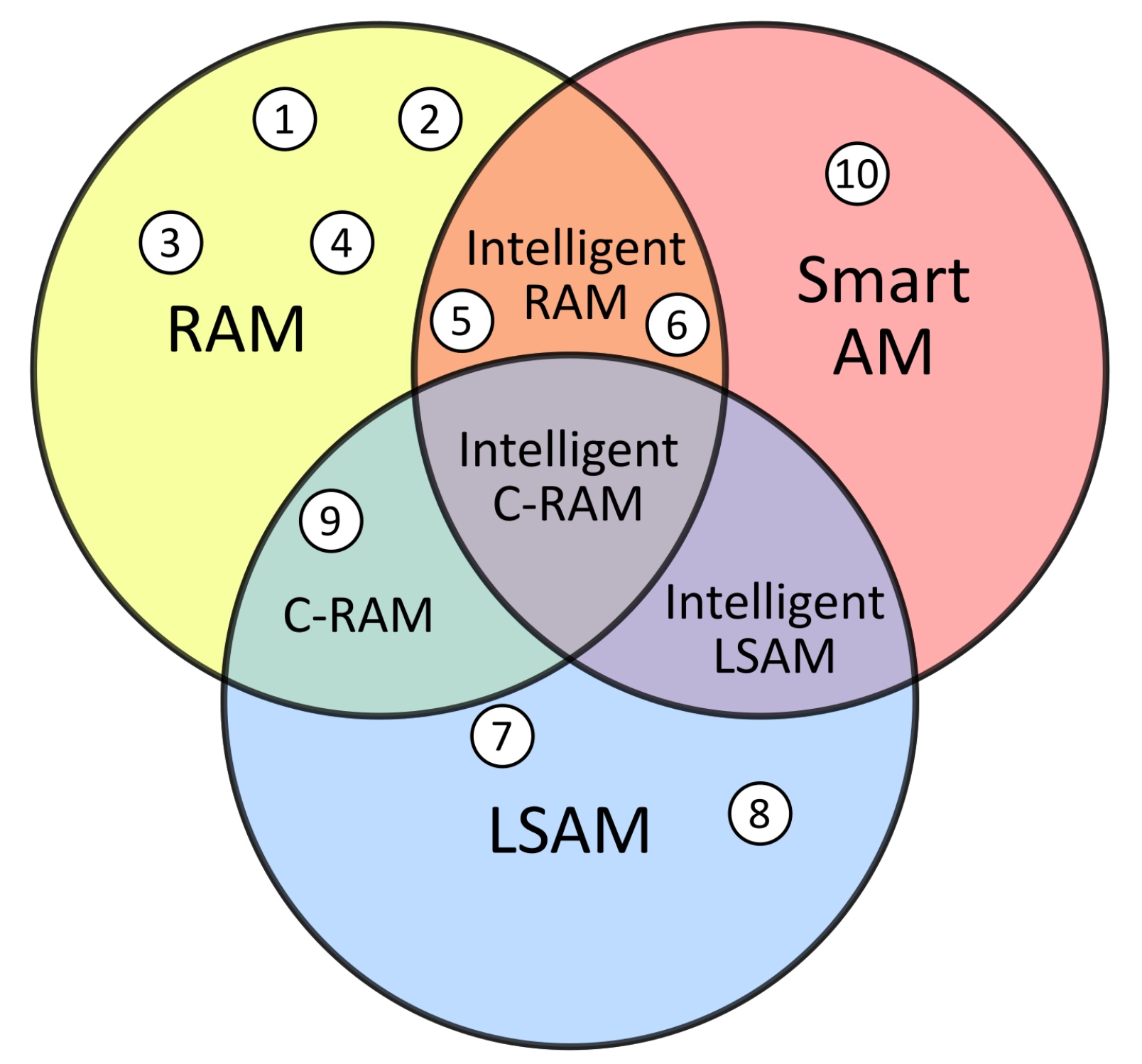} 
            \vspace{3mm}
            \caption{Relevant research areas of RAM and recent papers within these topics.}
            \label{fig:sample_figure}
        \end{minipage} &
        \begin{minipage}[t]{0.58\textwidth}
            \centering
            \vspace{-6.34cm}  
            \resizebox{\textwidth}{!}{%
            \begin{tabular}{m{.3cm}m{2.75cm}m{5cm}m{1.2cm}}
                \hline
                \textbf{No.} & \textbf{Author} & \textbf{Description} & \textbf{Area} \\
                \hline
                1 & Bhatt et al. \cite{bhattExpandingCapabilitiesAdditive2020} & Describes advantages of RAM systems including C-RAM & RAM \\

                2 & Jiang et al. \cite{jiang_review_2021} & Reviews high DoF gantry and RAM/C-RAM systems & RAM \\
                
                3 & Urhal et al. \cite{urhalRobotAssistedAdditive2019} & Reviews examples of RAM systems & RAM \\
                
                4 & Tang et al. \cite{tangReviewMultiaxisAdditive2024} & Covers advantages and challenges of multi-DoF systems & RAM \\
                
                5 & Zahid et al. \cite{zahidinResearchChallengesQuality2023} & Covers quality control strategies for WAAM RAM systems & Intelligent RAM \\
                
                6 & Chen et al. \cite{chenReviewWirearcAdditive2021} & Investigates defects and detection approaches for WAAM RAM systems & Intelligent RAM \\
               
                7 & Lehmann et al. \cite{lehmann_large-scale_2022} & Discusses state of the art large scale metal AM processes & LSAM \\
            
                8 & Vicente et al. \cite{vicenteLargeformatAdditiveManufacturing2023} & Review of large scale polymer material extrusion processes & LSAM \\
              
                9 & Alhijaily et al. \cite{alhijailyTeamsRobotsAdditive2023} & Discusses multi-robot AM systems & C-RAM \\
             
                10 & Xiong et al. \cite{xiong2022intelligent} & Reviews intelligent monitoring methods for AM & Smart AM \\
                \hline
            \end{tabular}
            }
            \label{tab:review_papers}
        \end{minipage}
    \end{tabular}
\end{figure*}

Cooperative robotic additive manufacturing (C-RAM) platforms comprised of multiple robotic arms enable improved fabrication speeds \cite{shen_research_2019}, enhanced sensing capabilities \cite{zimermannInprocessNondestructiveEvaluation2023}, and heterogeneous tooling for improved mechanical properties \cite{bhatt2019robotic}. As shown in Figure~\ref{tab:review_papers}, previous reviews have discussed the advantages of RAM systems \cite{bhattExpandingCapabilitiesAdditive2020, jiang_review_2021} and C-RAM systems \cite{alhijailyTeamsRobotsAdditive2023}, as well as improvements to metal RAM systems using machine learning (ML) \cite{heResearchApplicationArtificial2023}. 
These works do not comprehensively discuss the challenges of process quality and the relevant methods for obtaining desirable qualities for both single and multi-arm RAM systems. Furthermore, there have been a few discussions specifically regarding multi-arm RAM systems which are pertinent to resolving large-scale fabrication challenges in AM through the lens of part quality and mechanical properties.  

\begin{figure*}
    \centering
    \centerline{\includegraphics[scale=0.24]{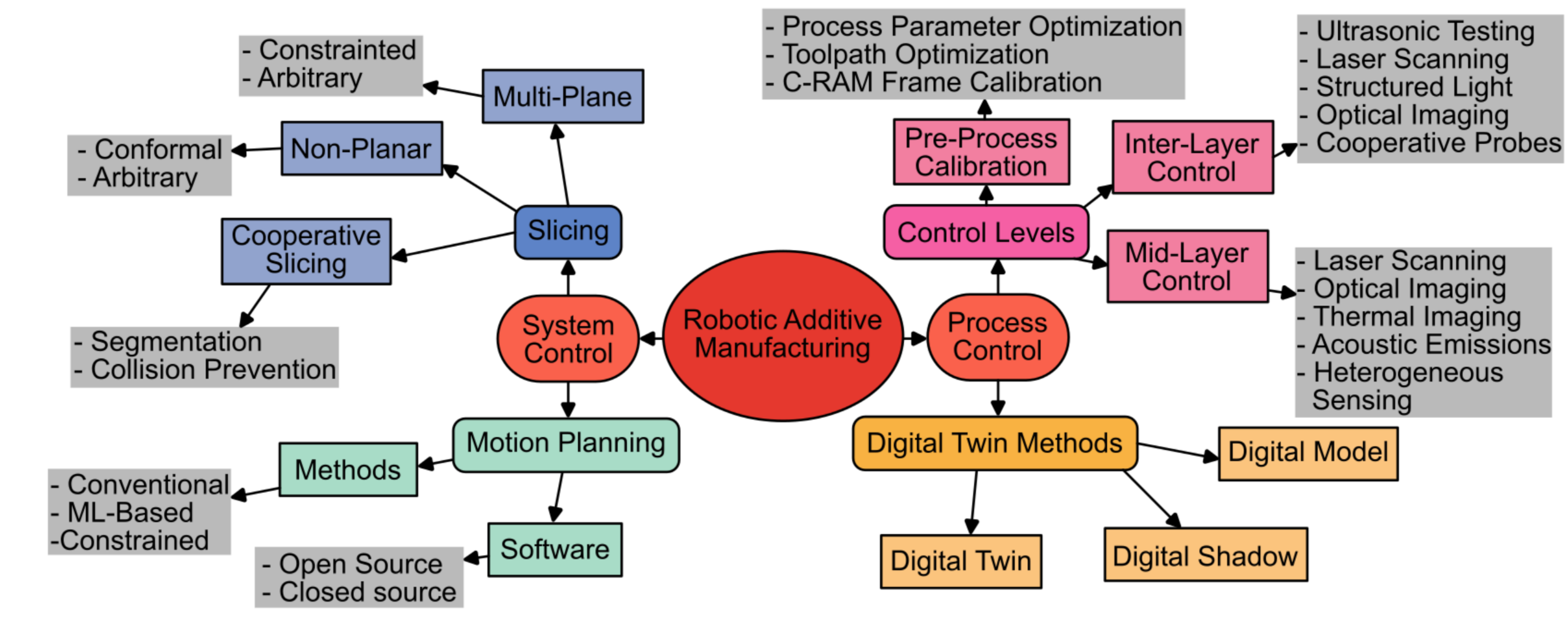}}
    \caption{Flowchart of the review paper which system control methods (i.e., slicing and toolpath planning) and process control methods for defect mitigation across process control levels (pre-process, inter-layer, and mid-layer) and using digital twins.}
    \label{fig:paper-structure}
\end{figure*}

In this paper, we review the aspects of C-RAM systems critical to intelligent and high-quality fabrication of parts (Figure \ref{fig:paper-structure}). An overview of the development of RAM and its advantages is introduced for both single and multi-arm systems. System control is then discussed regarding three main areas of focus, slicing (i.e. toolpath planning), motion planning, and digital twin. Process control methods for the mitigation of defects are introduced through three control levels. Challenges and opportunities within C-RAM systems as a reflection of the current literature are described preceding the conclusion of this work. The areas of investigation have considerable future impacts on emerging research fields such as digital twins which enable the harmonious integration of physical world feedback from sensors with digital representations. 
Specifically, this work provides the following contributions:
\begin{itemize}
\item C-RAM configurations are characterized by build volume into two categories: high-overlap (e.g., homogeneously tooled large-scale printing) and low-overlap (e.g., multi-material or cooperative sensing).
\item Toolpath planning (i.e., slicing) and motion planning (i.e., path planning and trajectory planning) methods are distinguished and discussed in depth in single and multi-arm systems. 
\item Defect mitigation techniques applicable to RAM systems are discussed, as well as challenges pertinent to C-RAM systems. Methods are characterized according to how feedback from these sensor signals can be employed: pre-process, inter-layer, or mid-layer.
\end{itemize}

The literature review conducted uses specific keywords to locate papers which utilize robotic systems for AM fabrication. As this is an emerging field with non-standardized terminology, a wider spread of terms is employed to locate works that fall under the definition of RAM. The search terms utilized to identify RAM papers include "robotic additive manufacturing", "robot|robotic", and "arm|manipulator", and for multi-arm RAM systems this adds "cooperative", "C3DP", "multi-arm", and "arms|manipulators".
Process-specific terms for RAM include "DED", "WAAM", "ME", and "FFF|FDM". 
Sensing specific terms for AM processes include "sensing", "quality", "monitoring", "detection", and "defect".
Combinations of these base search terms were used to locate the relevant papers covered within this review.

\section{Robotic Additive Manufacturing} \label{sec-ram-intro}

RAM systems are comprised of an AM process toolhead integrated into articulated robotic arms with $\geq$ 4-DoF (e.g., selective compliance assembly robot arm (SCARA), manipulator with spherical wrist).
RAM systems have out-of-bounds build volumes that are capable of being overlapped with additional systems or moved to a secondary location to enlarge the effective build volume and maximum part volume. In contrast, systems with in-bound build volumes (e.g., gantry variations) can only increase the maximum effective build volume through an increase in the size of the system itself.   
Parts are created through layer-by-layer deposition of material onto a substrate from an end effector integrated with a specific AM process (e.g., FFF, FDM, and DED). The framework by which parts are created is outlined as follows: A part created using computer aided design (CAD) is converted to a suitable file format (e.g., STL, OBJ) to then be processed by a slicing program. Pre-process operations such as segmentation or transformation are applied to the part according to the slicing method used. High degrees of freedom afford advanced path planning strategies such as multi-axis slicing, whereby additional axes of 2D planar slicing are defined, and non-planar slicing, where 3D surfaces are utilized to slice parts. 
The predetermined path of deposition, or toolpath plan, is then created using the desired slicing processes similar to conventional approaches \cite{zhao2020shape} to determine which points the end effector must move through to create the desired geometry.  
However, such high degrees of freedom found in RAM systems require sufficient inverse kinematic solutions, or motion plans, to determine what set of joint configurations should be used to move the end effector to the desired location and orientation \cite{liu2021review, tamizi_review_2023}. Solutions must be located which also avoid collisions and singularities which is further complicated in systems with multiple robotic arms.
Proper coordination of both the motion of the robotic arm and end effector control allows for the fabrication of parts with superior capabilities compared to conventional AM platforms.

Early examples of RAM systems focused on metal deposition processes which utilize modified welding processes and high DoF in robotic arms \cite{heralic2009towards}.
AM systems that can be compactly integrated into an end effector with moderate loads (<50kg) are most commonly used for RAM systems which include ME methods such as FFF or FDM, and DED methods such as WAAM \cite{bhattExpandingCapabilitiesAdditive2020}. 
One such early example is an investigation in 2011 by Bonaccorso et al. \cite{bonaccorso_arc_2011} with a modified welding process to enable layer-by-layer metal deposition implemented on a 6-DoF Kuka robot. 
Around this time RAM studies utilizing FFF/FDM processes were also investigated including a system proposed by Choi et al. \cite{choi2010virtual} in 2010 consisting of multiple SCARA robot arms on a modular plate.

\begin{figure*}
\centerline{\includegraphics[scale=0.39]{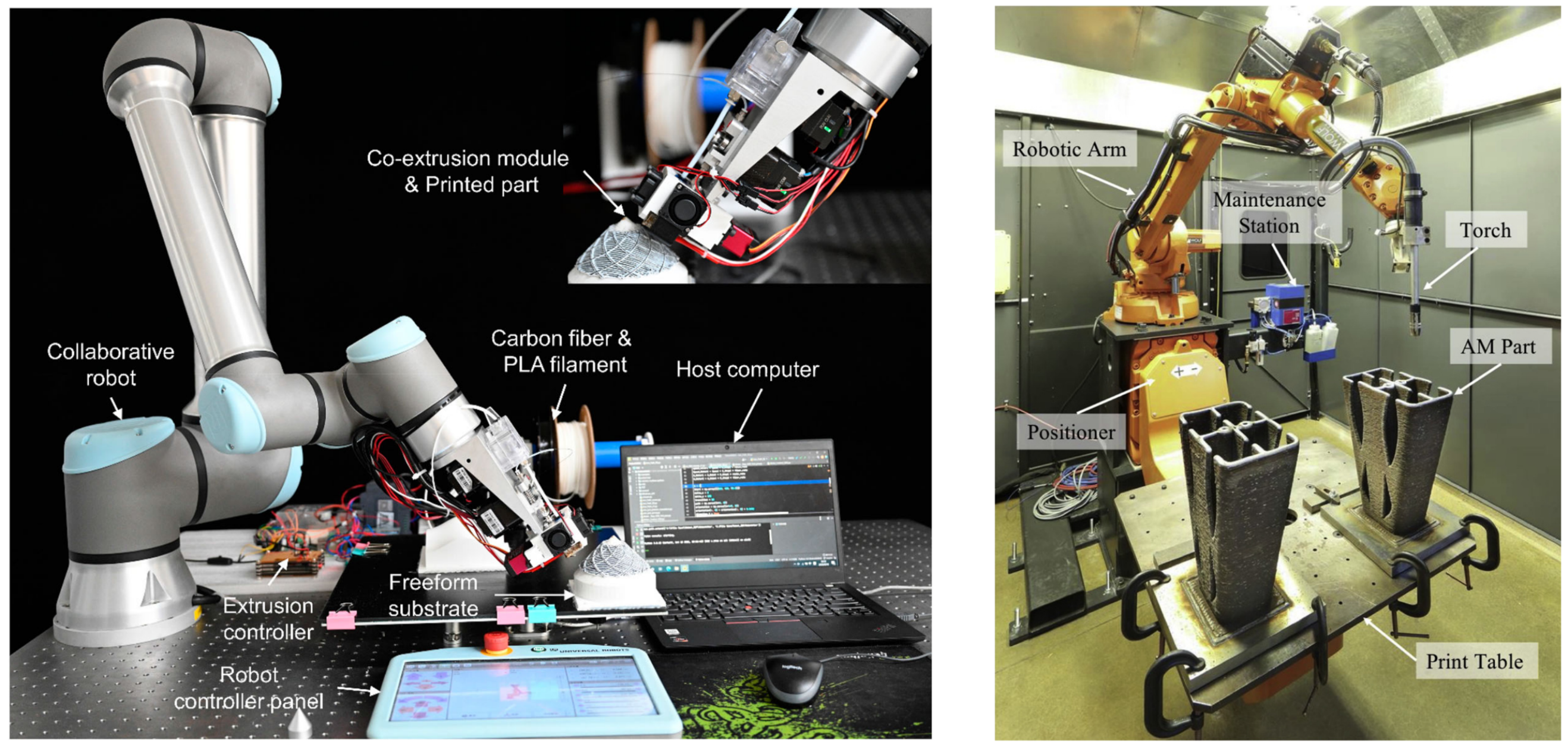}}
\caption{Single-arm RAM systems are typically utilized for material extrusion processes such as FFF \cite{zhang2023robot} (left) and directed energy deposition processes such as wire arc additive manufacturing \cite{hassenScalingMetalAdditive2020} (right).}
\label{fig:single-RAM}
\end{figure*}

Various robotic platforms have been used for RAM systems with differing DoF and subsequent capabilities \cite{jiangReviewMultipleDegrees2021}. For RAM systems, the most basic platform utilized in literature is 4-DoF robotic arms such as SCARA arms. These robotic platforms allow for large-scale and multi-arm systems but have limited uses in advanced slicing methods due to their relatively low degrees of freedom. The most common configuration of robotic arms used for RAM are 6-DoF systems which allow for motion in all directions within 3D space (Figure \ref{fig:single-RAM}). Such flexibility allows for the use of advanced slicing methods, with limitations resulting from the design of the end effector (e.g., maximum nozzle angle) and the AM process implemented. Beyond high DoF single arms, higher DoF systems come in two main forms: mobile platforms and rotary tables. To enable large-scale fabrication, authors have implemented high DoF robot arms onto mobile platforms for up to 3 additional DoF \cite{zhang2018large}. This allows for fabrication while the platform is moving or for the robotic arm to reposition to a new location which both increase the effective build volume of the system. Rotary tables are also commonly integrated into RAM systems which typically allow for the rotation of the part in additional axes. Notably, the use of redundant DoF allows for reconfiguration while retaining end effector orientation \cite{huber2021globally}. This allows for more flexibility for motion planning which can result in lower jerk and prevent the need for reconfiguration.

\subsection{C-RAM Systems}


\begin{figure*}
\centerline{\includegraphics[scale=0.45]{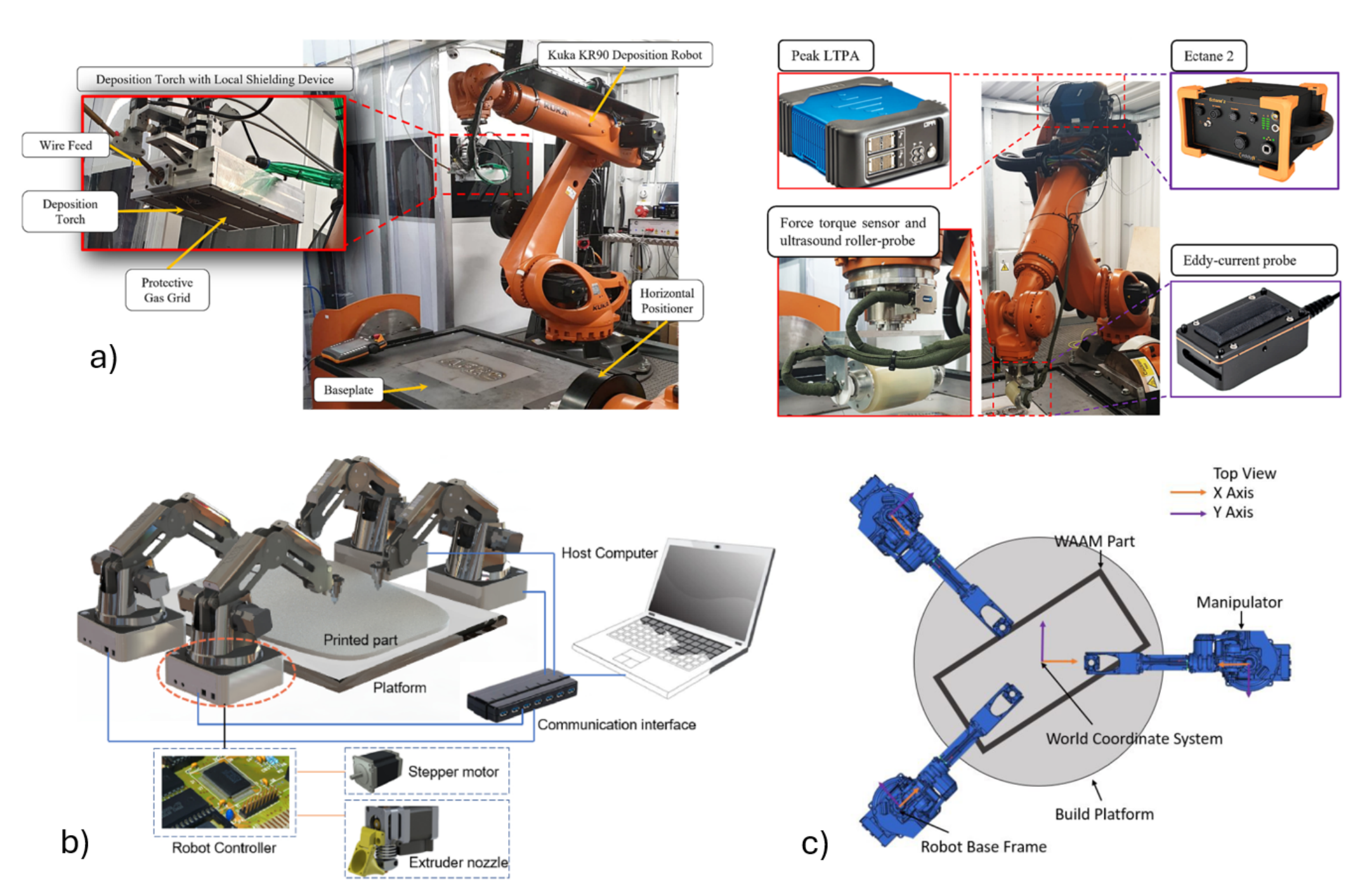}}
\caption{Multi-arm C-RAM systems offer varying capabilities according to their configuration including a) cooperative sensing capabilities \cite{zimermannInprocessNondestructiveEvaluation2023}, b) low overlap large scale printing \cite{shen_research_2019} and c) high overlap multi-arm fabrication \cite{bhattOptimizingMultiRobotPlacements2022}.}
\label{Fig-CRAM}
\end{figure*}

Multi-arm RAM systems, or C-RAM systems, utilize multiple robotic arms with overlapping build volumes to create components with enhanced capabilities. C-RAM configurations can afford improved fabrication speeds, increased maximum build volume, and heterogeneous tooling capabilities superior to that of single-arm RAM systems (Figure \ref{Fig-CRAM}). 
As these systems require the coordination of multiple arms to cooperate to complete a single part, they are indicated as cooperative robotic additive manufacturing systems whereas collaborative additive manufacturing systems \cite{xiong2023human} typically refer to coordination between robotic arms and humans.

\begin{figure*}
\centerline{\includegraphics[scale=0.35]{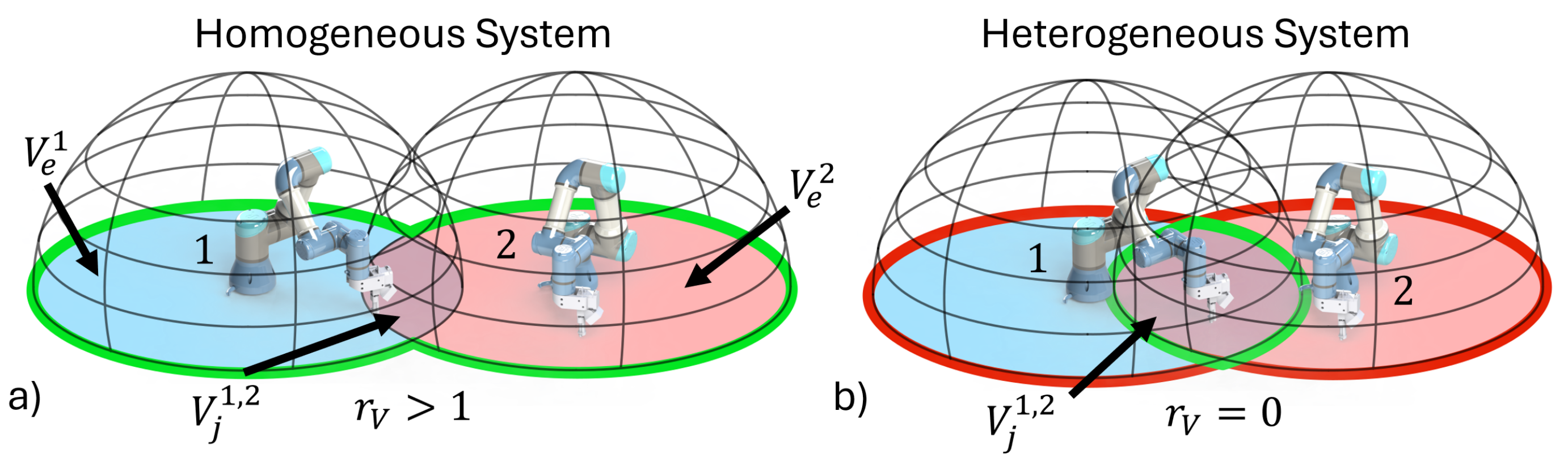}}
\caption{Multi-arm C-RAM configurations for a) low overlap (e.g., large-scale fabrication with homogeneous tooling) and b) high overlap (e.g., multi-material fabrication with heterogeneous tooling).}
\label{Fig-Overlap}
\end{figure*}

The capabilities of a C-RAM system are largely dictated by the usable build volume as defined by the placement and orientation of the integrated robotic arms. The critical regions within the reach of a robotic arm are fundamentally where material can be deposited without collision and onto a substrate which is a subset of the typical configuration space $C$ (i.e., the set of all valid arm configurations $\boldsymbol{q}$). From this subset of the configuration space, a single-arm RAM system has a build volume $V$ defined by the region in which the arm can deposit material given its geometric constraints. A C-RAM system which contains some $n$ number of robotic arms has build volumes $V_n$ of each robot arm in the system according to its configuration with the associated regions $V_{e(n)}$, the exclusive build volume which can only be reached by the $n$th arm, and $V_{j(n,m)}$, the joint build volume between overlapping arms $n$ and $m$ (Figure \ref{Fig-Overlap}). The effective build volume $V_{eff}$ is then the sum of all exclusive and joint build volumes of the system. The capabilities of a C-RAM system are well described by the proportion of the total exclusive build volume $V_e$ and total joint build volume $V_j$ defined by
\begin{equation}
    r_V = \frac{V_e}{V_j}
\end{equation}
A high-overlap C-RAM system possesses a volume ratio $r_V < 1$ where overlapping build volume exceeds exclusive build volume. C-RAM systems are high overlap most commonly in heterogeneous tooling configurations (e.g., multi-material, multi-resolution, and cooperative sensing) where multiple arms must work in or near the same region. A low-overlap C-RAM system possesses a volume ratio $r_V \ge 1$ where exclusive build volume is equal to or greater than overlapping. Such configurations are typically used in large-scale RAM applications where tooling is homogeneous; the joint build volume region is then necessary but not crucial to the capabilities of the system (and is often minimized). As all C-RAM systems must determine the location of placement for stationary arms, the selection of the value of $r_V$ by configuration must be carefully selected according to the desired capabilities of the C-RAM system.

Large-scale C-RAM systems which integrate multiple homogeneously tooled cooperating arms have been previously explored in both stationary and non-stationary configurations \cite{alhijaily2023teams}.
Shen et al. \cite{shen2019research} integrated four 4-DoF Dobot version 2.0 robotic arms for large-scale FDM fabrication through overlapping build volumes. Using this system a 300 mm $\times$ 250 mm part was fabricated with each arm completing a separate segment while avoiding collisions between arms. 
Such systems can be further expanded by additional arms, though this is primarily limited to a single axis as expansion in either the Z axis or in both X and Y axes are constrained by robot arm geometry and interference of the robot arm itself.
Mobile manipulators were used in coordination by Zhang et al. to fabricate a concrete bridge structure spanning over 2 meters in length \cite{zhang2018large}.
While mobile manipulators have greater advantages in expanding build volume, localization and coordination of arms within the system becomes increasingly complex as well as both path planning and motion planning of the system.
Both multi-arm approaches greatly improve the effective build volume of the system but incurr limitations resulting from the increased control complexity of the system.

C-RAM systems comprised of heterogeneously tooled arms offer alternative advantages and capabilities compared to homogeneously tooled systems. Bhatt et al. integrated two RAM systems with varying nozzle sizes (0.4mm and 0.8mm) to improve surface quality while retaining high fabrication speeds \cite{bhatt2019robotic}. Conformal printing was used for the secondary fine arm, further improving mechanical strength and removing the stepping effect typical of 2D planar slicing.
Secondary arms can also be used for multi-material printing for improved material properties. This includes fiber reinforced material extrusion where a secondary robotic arm lays fiber between layers to improve part strength \cite{fangExceptionalMechanicalPerformance2023}. 

\section{System Control for RAM} \label{sec-ram-system}
\begin{figure*}
\centerline{\includegraphics[scale=0.33]{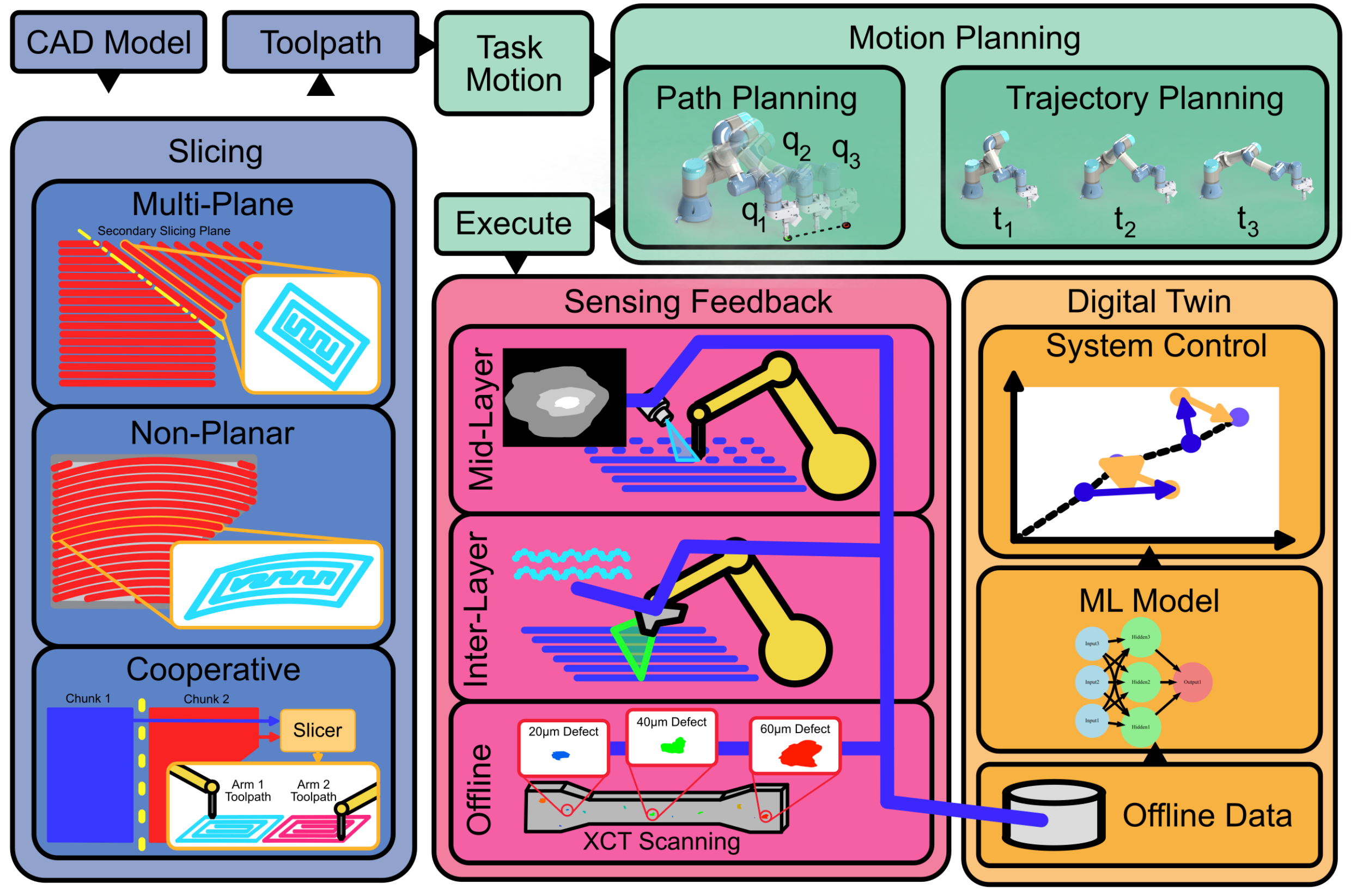}}
\caption{An example RAM workflow including slicing, motion planning, process control using sensing feedback, and digital twin models.}
\label{fig:slicing}
\end{figure*}

Robotic manipulators require sufficient control of motors to realize the desired motion and operation of the end effector. The planning and execution of this motion are conducted in two primary steps: toolpath planning and motion planning. Toolpath planning, or the determination of the desired path of motion of the tool center point (TCP), is defined according to the geometry of the part and is conducted before the motion planning of the arm. For RAM systems this is comprised of a "slicing" step whereby a part file is parsed into individual layers from which a toolpath can then be planned for each slice. Deposition of material following the toolpath then creates the desired geometry in a layer-by-layer fashion. Motion planning, or the determination of the inverse kinematic solution to afford motion to a desired point or end effector orientation, is conducted after a toolpath plan has been established to execute the desired motion of the end effector. For gantry-style AM platforms, motion planning is solved during the slicing step due to their trivial inverse kinematic solutions. High DoF systems with exceeding large potential configurations require motion planning to determine the path plan (the set of configurations to move between from the target and initial configuration) and trajectory plan (the time-dependent transition between path plan configurations). A general workflow of a RAM system and the role of slicing and motion planning is described in Figure \ref{fig:slicing}. The pertinent areas of research and methods applicable for RAM systems within slicing and motion planning are discussed in the following sections.

\subsection{Slicing}

Slicing for AM describes the multi-step process to generate instructional code from which a part can be fabricated in a layer-by-layer fashion. The code that is generated, typically g-code, is a function of various slicing parameters depending on the AM process (e.g., layer height, road width, and infill percentage). A 3D object file (e.g., STL and OBJ) is segmented into individual layers using surfaces where each layer is a set of closed loop contours. A toolpath plan is then generated from these contours to fabricate each corresponding layer; this can then be interpreted into the appropriate machine code (e.g. g-code) for the system. Commands that instruct arm motion can be task or transfer motions; task motions call for the deposition of material whereas transfer motions do not and are for repositioning (i.e., rapid commands). The determination of the toolpath plan within slicing is often described as path planning in RAM literature \cite{jiang2020path}. While this might be analogous to systems with trivial inverse kinematic solutions (e.g., 3-DoF gantry style printers), this is a misnomer for higher DoF RAM systems, as multiple solutions are possible and therefore only describes the motion of the TCP. To alleviate this ambiguity, the path generated during slicing for the end effector is specified as the toolpath.

Conventional slicing approaches are conducted during the pre-processing step where machine instructions are generated for the fabrication of the entire part in a one-time operation. Recently, authors such as Borish et al. have proposed the use of "on-demand" slicing where slicing is conducted in partial sub-sections for closed-loop feedback capabilities \cite{borish2021ornl}. Both open and closed loop slicing approaches follow the same general slicing steps, with the main distinction of closed-loop being the segmentation of the CAD file and using sensing feedback to modify the slicing process in-situ.

\begin{figure*}
\centerline{\includegraphics[scale=0.45]{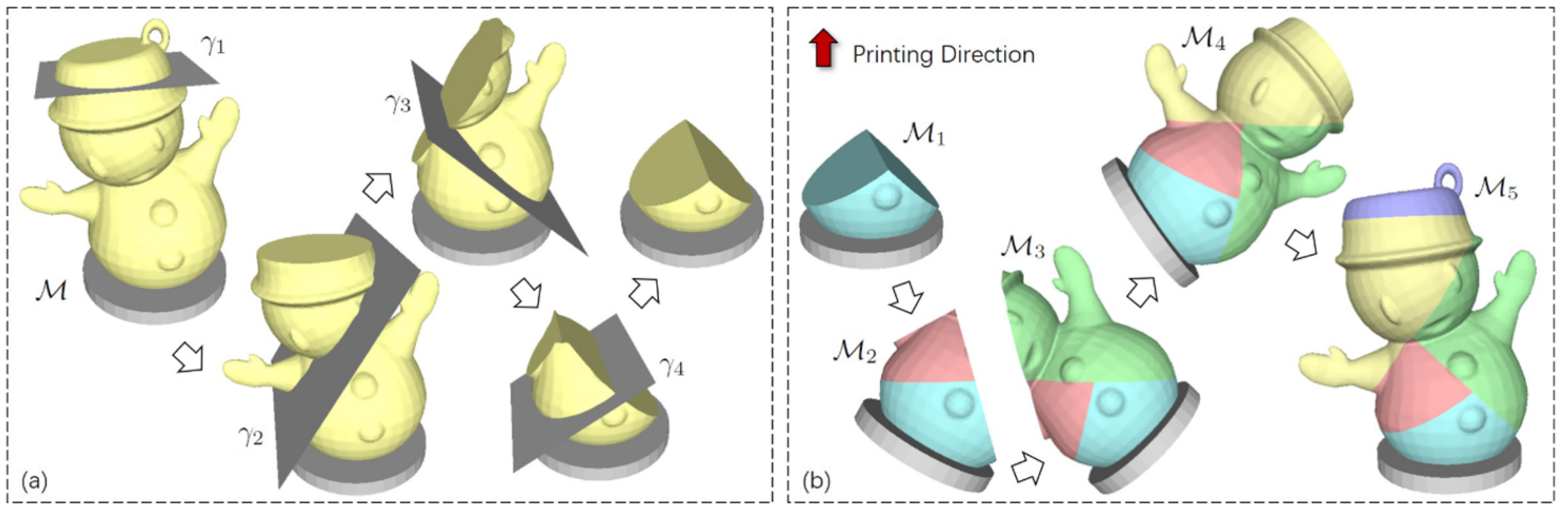}}
\caption{Multi-plane slicing method proposed by Wu et al. \cite{wu2019general} comprises of a) decomposition to eliminate material usage which is then translated into b) ordinal multi-plane slicing segments.}
\label{Fig-Multi-Plane}
\end{figure*}

Conventional AM systems utilize 2D slicing procedures whereby a single axis is used as a reference to then generate 2D cross-sectional slices according to a desired layer height. More slicing methods which modulate the shape and orientation of these layers are enabled from higher DoFs typical of RAM systems affording reductions in support material usage and enhanced part strength. For C-RAM systems that integrate multiple RAM systems, the decomposition of parts into sub-sections must be determined. These sections then undergo typical slicing processes to enable parallel cooperative printing and realization of the final desired geometry. The following section reviews advanced approaches to slicing (i.e., multi-axis and non-planar) for RAM systems in addition to decomposition methods and considerations for C-RAM systems.

\subsubsection{Multi-Plane Slicing}

While conventional 2D planar slicing uses a single plane from which slices are created, multi-plane slicing defines multiple planes for reduced support material usage and improved part strength. Auxiliary planes are created according to user criteria or through part decomposition algorithms which decompose parts to optimize objective functions. Various approaches have been investigated for multi-plane slicing primarily in FFF processes due to their reliance on support material for complex geometries and overhangs. Approaches for multi-plane slicing fall into two categories, discrete and continuous plane shifting.

Discrete axis shifting uses a set of predefined planes to switch to which are typically selected according to geometrical constraints such as the fabrication of lattices.  Ishak et al. proposed a discrete multi-plane RAM implementation for FFF process integrated with a 6-DoF Motoman SV3X robotic arm \cite{binishakRobotArmPlatform2016a, ishakRobotArmPlatform2016}. For specialized geometries such as lattice structures, multi-plane slicing can reduce the amount of layers and subsequent inter-layer seams to improve mechanical strength \cite{ishakRobotArmPlatform2017}. To create these lattice structures, the original STL was first decomposed using a custom Matlab script into individual segments according to the plane from which they were sliced. These individual sections were then sliced in a conventional 2D planar manner using Repetier software in the default normal plane orientation; the subsequent g-code was then transformed back to the desired plane orientation with respect to global coordinates and recomposed into a single tool path code. High-quality fabrication of lattice structures in 4 different planes was achieved with this method with structure sizes corresponding to the size of the nozzle used (maximum 2mm) \cite{ishakMotoMakerRobotFDM2019}. It is unclear how well the lattice toolpath method generalizes to more complex lattice geometries compared to the structures utilized to evaluate the approach, as well as the exact mechanical performance increase that can be observed by utilizing this kind of technique. The greatest benefit of discrete axis shifting is in processes where intralayer bond is a constraint that can be mitigated which is an important consideration for FFF, while other AM methods stand less to gain from this path planning approach.

Continuous axis shifting utilizes planes generated at arbitrary orientations to reduce support material usage which subsequently improves surface quality, and reduces waste material as well as post-processing operations. Wu et al. proposed a decomposition algorithm for continuous multi-axis path planning comprising of shape-based coarse decomposition, sequence planning, and constrained fine tuning \cite{wu2017robofdm}. The methodology was implemented on a UR3 robotic arm controlled using robotic operating system (ROS) with a stationary extruder printing on a build plate mounted on the end effector of the arm, allowing for support-free printing of various geometries. This approach was further extended by Gao et al. \cite{gao_near_2019} which implemented global optimization criteria to locate the planes which minimize the support material area. Wu extended his original algorithm using various optimization criteria including greedy, constrained greedy, and beam-guided greedy schemes aiming to determine the best decomposition strategy depending on part geometry \cite{wu2019general}. These implementations each displayed methods of greatly reducing or totally removing support material usage, with criteria-based methods showing the best performance in locating optimal part decomposition strategies. However, these implementations rely on the readjustment of the build plate to be normal to the direction of gravity to eliminate the effect of overhangs while the extruder is in a stationary configuration which limits the effective build volume of the system. In instances where a substrate for subsequent planes is not built from previous layers such as examples explored by Insero et al., layers must utilize variable layer heights to reduce uneven material deposition. In this work a variable infill algorithm is introduced for both planar multi-plane \cite{insero_novel_2022} and later non-planar \cite{insero_non-planar_2023} multi-plane printing to reduce layer thickness variation during printing. As noted by the authors, the parts used to evaluate this method still observed undesirable layer variation in addition to not being suitable for geometries such as saddle surfaces.

\subsubsection{Non-Planar Slicing}

\begin{figure*}
\centerline{\includegraphics[scale=0.45]{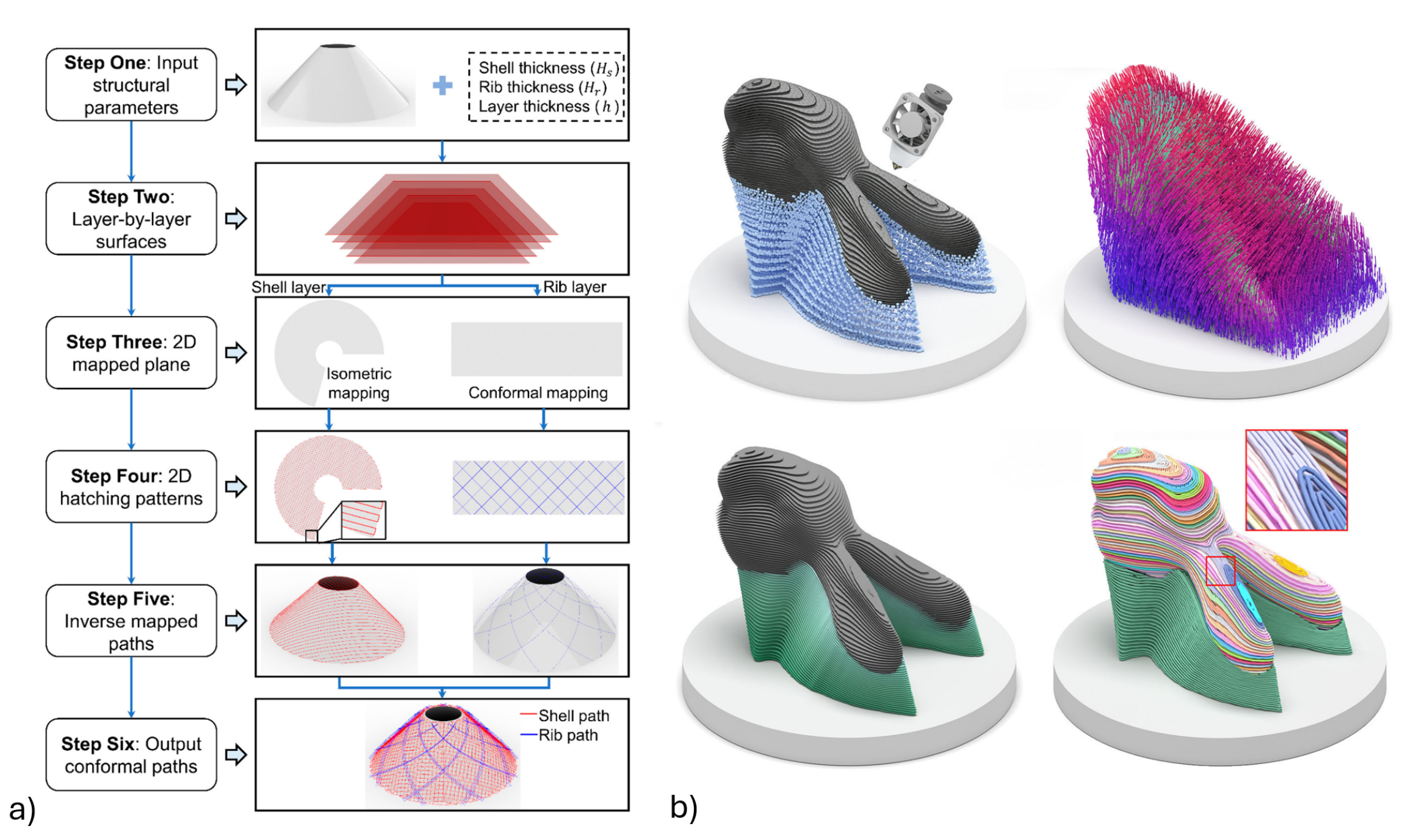}}
\caption{Non-planar slicing falls into two categories, a) conformal which uses a top surface as a reference \cite{zhang2023robot} and b) arbitrary which uses alternative metrics such as a heuristic to define slicing planes \cite{fang_reinforced_2020}.}
\label{Fig-NonPlanar}
\end{figure*}

Non-planar slicing methods slice parts according to 3D surfaces rather than 2D planes as conventionally used. Non-planar slicing enables the reduction or removal of support material for overhang geometries, enhances surface roughness, and improves material strength for materials such as reinforced carbon fiber \cite{bhatt2019robotic}. Non-planar slicing for gantry style platforms has been investigated for over a decade \cite{adams2011conformal} using both 3D toolpaths and the use of rotary tables, while high DoF AM and RAM systems have seen more recent use \cite{nayyeriPlanarNonplanarSlicing2022}. The maximum convex angle is limited by the geometry of the end effector and the flexibility of the system, lending greater advantages of non-planar slicing to RAM systems over gantry-style platforms. Non-planar slicing methods are categorized in two primary approaches: conformal and arbitrary. 

Conformal non-planar slicing methods use a top reference surface to slice parts which improves surface finish by removing the stepping effect observed with conventional 2D planar slicing. Moreover, the smoothed top surface creates enhanced part strength \cite{yao_3d_2021} which can be further increased through the use of fiber reinforced materials \cite{bhatt2019robotic, zhang_robot-assisted_2023}.  Applications of conformal slicing for RAM have been investigated in FFF processes as early as 2015 \cite{zhang_robotic_2015} with many examples while applications with DED processes have only more recently been investigated. The most common means of generating conformal toolpaths is by projecting a planar toolpath onto a reference surface, creating a 3D conformal toolpath that can then be registered to the top layer of a part \cite{shembekar_generating_2019, zhang_robot-assisted_2023}. While sufficient for creating initial conformal toolpaths, further adjustments must be made to ensure sufficient print quality. Bhatt et al. explored applications of conformal slicing in DED processes with non-planar substrates in addition to modifying tool paths to reduce material over-deposition in sharp corners and ensure proper material deposition \cite{bhatt_automated_2022}. Laser scanning was later used to evaluate the geometric deviation found between parts for post-processing machining operations \cite{bhatt_robot_2022}. 
Conformal non-planar slicing can also afford the creation of overhang structures that would not be suitable for conventional slicing methods. Kaji et al. \cite{kaji_process_2022} created a hollow thin-wall dome using a 6+2-DoF L-DED RAM system which used a custom slicing algorithm to generate toolpaths which avoids collision between the nozzle and the part according to the nozzle geometry and deposition overhang. A maximum overhang of 32.5\ang{30} was achieved with 2\% diameter deviation of the finished part. The approach relies on the use of a 2-axis servo positioner, however, the algorithm's applications to conventional 6-DoF systems are unclear.
Multiple reference surfaces can be used to enable smooth surfaces on both the top and bottom surfaces of prints as explored by previous authors in FFF processes \cite{fortunato_fully_2023, bhatt2019robotic}. The application of these kinds of conformal toolpath strategies which rely primarily on projection have limited application to complex geometries. Further development of algorithms to handle complex surfaces (e.g., toroidal) must be realized to leverage conformal benefits across a wider range of part geometries. 

Arbitrary non-planar slicing methods define a 3D surface from which slicing is conducted that is not defined by a reference surface of the part. For FFF processes which use fiber reinforcement this can allow for the careful design on reinforcement regions for substantial improvement of mechanical properties. Arbitrary slicing methods commonly use affine and non-affine transformations for regions to then be sliced in a planar manner; these sliced paths then undergo an inverse transformation to regain the original shape with a transformed non-planar toolpath plan. The basis for the transformation is typically defined by the user or according to the coordinate system utilized (e.g., spherical, non-planar).

Coordinate system shifts for improved fabrication of parts according to their geometry have been commonly explored by previous authors.
Zhao et al. developed a non-planar slicing method to reduce the reliance of support material for parts such as propellers through a cylindrical slicing approach \cite{zhao2018nonplanar}. The propeller features are separated from the main body of the cylinder of the propeller to allow for individual slicing. A transformation inverse to the desired slicing geometry is applied to the individual segments which then is sliced in a planar manner. The slice then undergoes the reverse transformation to regain the original geometry and create curved slices. This methodology is also applied to a curved part fabricated on a curved build plate. 
Geometries with overhangs pose a problem for DED processes, for which Dai et al. developed a novel PCA-based toolpath planning algorithm to create additional support for subsequent overhang layers \cite{dai_multiaxis_2022}. 
Mechanical strength of FFF parts that use continuous fiber reinforcement can be greatly enhanced using arbitrary non-planar slicing. Continuous fiber-reinforced thermoplastic composites (CFRTPCs) utilize in-nozzle or out-of-nozzle \cite{tao2023review} impregnation of fibers to enhance mechanical strength axially along fiber orientation over conventional FFF processes. Therefore, slicing methods which optimally orient fibers for enhanced mechanical properties can fully leverage the advantages of CFRTPC processes.  
Fang et al. \cite{fang_reinforced_2020} developed a slicing method to generate toolpaths for a cooperative fabrication RAM system to optimally orient fibers according to the stress fields of printed components. Stress fields of the desired component are created from finite element analysis (FEA) from which principal stress lines are generated. The set of principal stress lines then guide the arbitrary non-planar slicing process in which continuous fiber reinforcements are placed in a similar manner. Additional optimization is conducted to ensure continuity around critical load-bearing regions to maximize strengthening from reinforcement. Failure loads and model stiffness of the case study components observe improvements ranging from 105.1\%-544.0\% and 59.5\%-140.2\% improvements, respectively. The authors note high fabrication times due to slow feed rates of robot motion and limits in their optimization approach to remove small slice patches. Furthermore, the bracket-type parts evaluated share some homotopy-equivalence and are evaluated with simple (single) forces; the capabilities of the methodology to handle higher complexity parts and loading which result in more complex principal stresses is unclear.
The advantages of non-planar CFRTPC printing were also explored by Zhang et al. for the printing of reinforced shell structures \cite{zhang_robot-assisted_2023}. A workflow for high DoF CFRTPC printing using a 6-DoF UR3 robotic arm is proposed and evaluated on a conical shell structure. Greatly improved compressive strength and stiffness were observed by properly orienting the fiber directly versus conventional planar printing methods. 
Support of arbitrary non-planar layers prevents the use of conventional support algorithms used in FFF software inspiring Zhang et al. \cite{zhang_support_2023} to develop a support generation algorithm to address this drawback. A conventional support-tree algorithm is extended to handle instances where nodes must be joined from differing layers of deposition typical of non-planar printing. According to threshold angles defined, nodes are used to generate the support-tree join such that non-planarly sliced prints can be suitably supported for high print quality.

\subsubsection{Considerations for C-RAM}
For C-RAM systems, which implement multiple extruder robotic arms, a part must be decomposed into separate sections such that each arm can contribute to the generation of the final geometry in parallel. Fabrication must occur while avoiding collision between arms which can be accomplished during toolpath planning or during motion planning. The following sections will discuss segmentation and pre-process collision avoidance approaches for C-RAM systems. 

\begin{figure*}
\centerline{\includegraphics[scale=0.45]{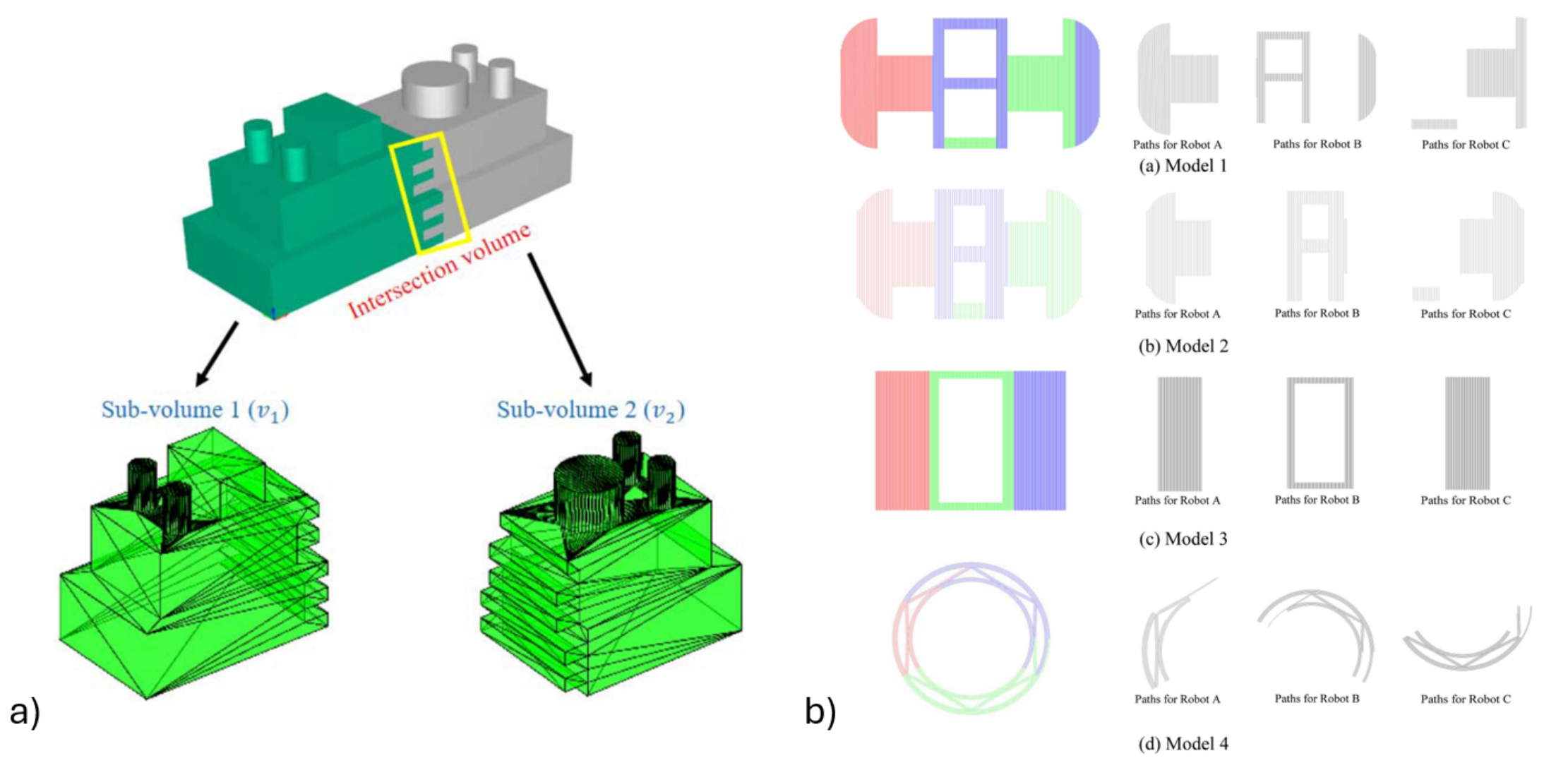}}
\caption{Segmentation for C-RAM systems defined by a) an intersection surface for two arms \cite{manoharan_novel_2020} and b) planar segments for three arms cooperative printing \cite{li2024allocation}.}
\label{Fig-Segmentation}
\end{figure*}

\textbf{Segmentation:} Decomposition of CAD models into separate segments is a critical requirement of all C-RAM systems which utilize build regions that exceed the intersection region between build volumes, meaning areas of the part can only be created by a subset of arms within the system. Therefore there will always be some intersection or boundary between parts fabricated in complementary regions that then meet sections located within intersection regions. Furthermore, part segmentation is necessary to allow for arms to work in parallel in any C-RAM configuration. The determination of these sub-sections of parts is therefore critical to the structure of a C-RAM system.

C-RAM systems whose configuration has little overlap between build volumes, typically for large-scale applications, have subsequently small regions from which the section boundary can be defined. Nevertheless, the selection of boundary location to create sections and its interface surface plays a critical role in the quality of the final component and its mechanical properties. As a part of a novel FFF C-RAM system developed by Shen et al., a segmentation algorithm is introduced to minimize the difference in layer completion between each fabrication robotic arm \cite{shen2019research}. Simulated annealing is utilized with an objective function that seeks to evenly distribute the internal region of the contour. A relaxation technique developed by Metropolis et al. \cite{metropolis1953equation} is used to prevent local optimal solutions. The resulting decomposition is sufficient to ensure near-equal layer competition times, but is dependent on the position of the part with respect to the build volume (e.g., center of the part and part geometry) in addition to being implemented for processes which must maintain equal layer heights during fabrication. In instances where a C-RAM system can allow for desynchronized layer fabrication, including instances of advanced slicing methods, such an approach would not be required. While most methods of part decomposition utilize 2D planes to split parts, recent works have investigated alternative 3D surfaces by which sections can be defined. 
Manoharn et al. \cite{manoharan_novel_2020} in a preliminary work proposed a methodology to create components using a C-RAM system with an interlocking interface structure to increase bond strength between decomposed volumes. 
Similarly, Stone et al. \cite{stone2023print} utilized non-planar interface layers to define decomposition regions to evaluate the performance of a novel mid-process collision avoidance algorithm.
These studies do not investigate the impact of these interface patterns on mechanical properties as their focus is on system development improvements (i.e., decomposition and collision avoidance improvements, respectively). As FFF processes suffer from inter-layer weakness which gives rise to its antistropic properties \cite{gaoFusedFilamentFabrication2021b}, it is critical to the feasibility of C-RAM systems to understand how the definition of the inter-section boundary affects the overall mechanical properties of parts. DED similarly suffers from antistropic properties \cite{wolff2016anisotropic} again suggesting that the selection of interface orientation and shape has an impact on mechanical properties.


Segmentation for C-RAM can also be conducted on a discrete point basis rather than entire sub-sections of the part.
Arbogast et al. developed a bead scoring system for discrete deposition points within a C-RAM DED process to determine the ordered list of beads to assign to each of the three robots within the cooperative system
\cite{arbogastStrategiesScalableMultirobot2024}. Three ABB IRB 4600 robots were placed such that their build volumes each overlapped a center positioner table resulting in a high-overlap C-RAM system. Points within a layer which are closer to the base of the robot score higher helping to ensure that arms fabricate close to their base and away from other arms. In addition, beads were marked as either local or global depending on the distance of the bead from the center of the base of any arm; if the bead stretches out across the entire build plate and requires high clearance from other arms, it is considered a global bead. The printing process is then controlled such that all arms work jointly on either local or global bead regions. The proposed methodology is sufficient for high overlap C-RAM but it is not clear how it generalizes to low overlap configurations. In addition, there is no discussion relating to the collision avoidance method used after assigning beads to robotic arms which is not guaranteed from the existence of multiple valid inverse kinematic solutions.

\textbf{Collision Prevention:} In C-RAM systems where collision is possible between arms during fabrication, collision avoidance can be assured through toolpath planning considerations in a preventative pre-process step through the determination of toolpath order. Depending on the system configuration, such as a low-overlap C-RAM platform, scheduling algorithms can be used to prevent the opportunity of collision during printing without the need for in-situ collision assessment \cite{shen2019research, arbogastStrategiesScalableMultirobot2024}. Regions or operations where collisions could take place (e.g., printing in overlap build regions or overlapping operations) are identified during the slicing process; operations are then ordered such that multiple robots do not work in the marked region at the same time, preventing the potential of collisions. This inter-robot collision avoidance approach is implicit and applicable to simple system designs and part geometries where overlap is low and section borders are trivial (i.e. planar, high separation). For higher complexity systems or parts within high build volume overlap, implementing such an approach would result in high system constraints and extended wait periods to prevent building within overlapping build areas. In these instances, a combined toolpath and motion-planning-based collision avoidance approach that affords collaborative fabrication within overlap regions is more desirable.

\subsection{Motion Planning}

\begin{figure}
\centerline{\includegraphics[scale=0.23]{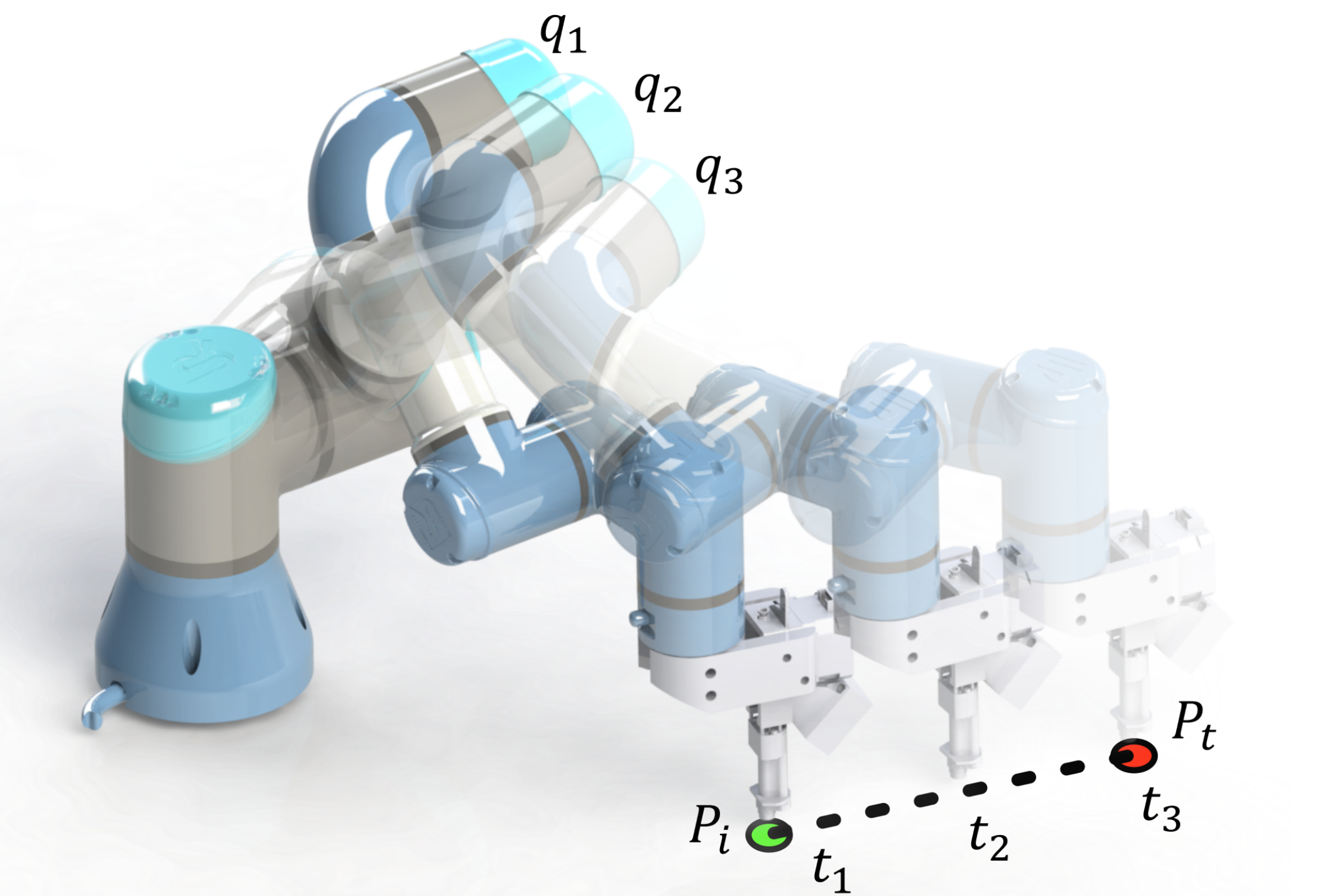}}
\caption{Motion planning structure which includes path planning and trajectory planning.}
\label{Fig-MotionPlanning}
\end{figure}


Once the required toolpath has been established through slicing, the control of the robotic system to achieve the desired motion must be determined. The machine commands for task or transfer motion begin at an initial position $q_{initial}$ and specify a final position $q_{final}$ where $q = [\theta_{1}, \cdots, \theta_{n}]$ for an $n$ DoF robotic system. The configuration $q$ is an element of the configuration space $C$ from which we must locate a valid solution which is obtained via motion planning. Motion planning is comprised of two aspects, path planning and trajectory planning \cite{tamizi_review_2023}. Path planning seeks to determine the set of configurations $\boldsymbol{q}$ to transition between the initial configuration $q_{initial}$ and target configuration $q_{target}$ \cite{latombe2012robot}. Shifting between the ordered set of intermediary configurations through the control of joints allows for control of the target configuration. Trajectory planning defines the time-dependent transition between the set of configurations to obtain joint velocities $\dot{q}$ and accelerations $\ddot{q}$.  The motion plan solution must take into account the geometric and kinodynamic constraints of the robotic system \cite{pham2015trajectory} in addition to task constraints (i.e. task or transfer motion) to obtain valid solutions. Motion planning methods for RAM must effectively locate optimal solutions for both path planning of intermediary points and the associated trajectory plan.

 As RAM systems operate within a constrained setting where a part is being fabricated, the optimal motion planning solution must also avoid collision with the surrounding environment \cite{tamizi_review_2023}. Therefore, valid solutions of the calculated motion plan must be located in the subset of the configuration space $C$ which does not collide with obstacles $C_{free}$ with all other non-valid solutions which do collide being under subspace $C_{obstacles}$  \cite{choset2005principles}. Distance measurement algorithms for triangular meshes, such as the Gilbert-John-Keerthi (GJK) algorithm \cite{gilbert1988fast}, are used to determine collisions with spherical approximations of objects commonly used to reduce computational complexity \cite{hubbard1996approximating}. The determination of $C_{obstacles}$ accounts for the majority of path planning time in conventional methods \cite{kleinbort2020collision} as a result of the number of queries required to locate a solution. Therefore, much of recent research has sought to improve the search of $C$-space through the use of ML methods \cite{das2020learning, pan2016fast}, though these approaches are not well suited for dynamic environments. 

Motion planning for C-RAM systems must also consider the potential for inter-arm collisions which can occur when both arms are operating in overlapping build volume regions.
Stone et al. developed an in-situ collision detection mechanism whereby a pause would be instantiated if the next motion would result in an overlapping collision region for SCARA arms for low-overlap C-RAM \cite{stone2023print}. When the next position is determined from the sliced code during motion planning, the area created by the 2D orientation of the SCARA arms between its original and final position is calculated with respect to each robot's location. The areas are then checked to see if they overlap; if the overlapping condition is met, a potential collision has been detected in the next action. One arm is then selected to pause for that operation, and continues when the check passes again. The algorithm used is lightweight and allows for this estimate in real time during the motion planning phase. However, the method takes advantage of the geometry of the SCARA robots which allows for easy area estimation. The calculation for the area becomes more complex as the degree of freedom increases, therefore it is unclear how much of a computational burden and delay is imposed as system complexity elevates. Methods for motion planning-based collision detection must be lightweight enough to calculate in inter-layer or pre-process steps to ensure minimal delays are imposed.

Motion planning can be conducted in an offline or online manner depending on the computation efficiency of the approach and the complexity of the motion planning problem. For repetitive tasks such as welding or production, offline planning is generally suitable. The same can be said for RAM applications which use feed-forward control, but has limitations in adaptability which is critical for defect mitigation. In such instances where sensing feedback can be leveraged, online approaches are more appropriate as the toolpath plan, and subsequently the motion plan, then can be changed on the fly with minimum delay. As sensing is an integral aspect to maintaining quality of AM processes, it follows that online motion planning approaches are most suitable for RAM applications where mechanical properties of components are a primary concern.

\subsubsection{Motion Planning Methods}
Various methods have been explored in the literature to optimally locate motion planning solutions for both path planning and trajectory planning \cite{guo_recent_2023}. For RAM applications the toolpath (i.e. the path of the end effector) task motions are predefined during the slicing step which constrains the motion planning problem compared to transfer motion applications (e.g., pick and place, assembly). Furthermore, optimal solutions for task motions do not consider the minimization of execution time, as the velocity profile of task execution is predefined. Selection of motion planning methods therefore must optimally locate valid path plans and trajectory plans that minimize defect generation and minimize motion planning time. 

Particular considerations must be made for the selection of motion planning methods for use in RAM systems as path planning and trajectory planning outcomes have a massive impact on the quality of the AM process.
Motion planning conventionally seeks optimal solutions according to cost functions which aim to minimize criteria such as computational time, path length, and path jerk \cite{lynch_modern_2017}; this typically is in the context of transfer motions and can result in non-linear path planning between initial and target configurations. For RAM systems that predefine the intermediary path of the TCP, minimization of jerk and computing time are paramount whereas the TCP trajectory between paths is determined by the task. Motion planning algorithms which efficiently search the configuration space given these constraints are most suitable for RAM applications.

Conventional motion planning algorithms for industrial robotic arms primarily fall into three main categories \cite{tamizi_review_2023}: artificial potential field (APF), bio-inspired heuristic, and sampling-based methods. APF was first introduced by Khatib et al. \cite{khatib1986real} where obstacles apply an artificial repulsion force and the desired final location applies an artificial attraction force such that the total artificial field is differentiable to locate a solution. Its primary advantage lies in dynamic environment applications \cite{tamizi_review_2023}, but suffers from local minimums which prevent solutions from being found \cite{gasparetto2015path}. Various works have investigated approaches to minimize the local minimum drawback but not without additional drawbacks such as suboptimal solutions or increased complexity \cite{tamizi_review_2023}. 
Bio-inspired heuristic methods include approaches such as genetic algorithms \cite{ahuactzin1993using}, particle swarm optimization \cite{kim2015trajectory}, and ant colony optimization \cite{hamdoun2016optimal}. One of the most widely used heuristic methods for motion planning is the A* algorithm \cite{guruji2016time} that evaluates heuristics on nearby nodes to locate optimal paths. Variations of A* include Theta* \cite{nash2007theta}, which propagates information along edges, and D* \cite{stentz1995focussed}, a dynamic variation that is able to handle changing environments. While heuristic methods have good overall performance they generally require long execution times to determine optimal paths in complicated environments \cite{tamizi_review_2023}.
Sampling-based methods include probabilistic roadmap method (PRM) and rapidly-exploring random trees (RRT). PRM locates an optimal path in a discretized representation of the configuration space \cite{kavraki1996probabilistic} in a probabilistically complete manner \cite{geraerts2004comparative}. RRT creates random trees within the configuration space that are incrementally expanded to random nodes within the c-space \cite{lavalle1998rapidly}. The generation of random nodes has a large impact on the performance of RRT leading to variations which use biased sampling to minimize exploration \cite{qureshi2013adaptive, gammell2015batch}.
These motion planning algorithms are most widely developed to locate optimal solutions for transfer motions with unconstrained TCP paths. RAM applications are comprised of primarily task motions and constrained TCP paths which greatly reduces the complexity of locating valid motion planning solutions. For these reasons commonly used methods such as A* or RRT methods suffice, though they can be inefficient in location solutions due to not properly handling motion constraints.

ML motion planning methods can offer superior motion planning solutions to conventional methods.
Deep learning (DL) is one of the widely used ML models due to its capabilities in complex contexts over basic neural network architectures \cite{shrestha2019review}. DL is most commonly used to improve the random sample generation of sampling-based motion planning methods to improve efficiency \cite{tamizi_review_2023, ying2021deep}. DL has also been applied as a standalone motion planning network (MPN) by Qureshi et al. comprised of an environment encoding model and a planning model which showed efficient performance in up to 7-DoF systems \cite{qureshi2019motion}. DL methods have also been applied to constrained motion planning applications by Qureshi et al. as an extension of the MPN called constrained motion planning network (CoMPN). Model inputs are extended to include start and goal constraints to create motion plans which follow constraint manifolds defined by the task. The model was further improved to decrease computational time by supplying sampling-based methods with efficient sampling nodes \cite{qureshi2021constrained}.
Reinforcement learning (RL) methods have also been widely investigated which use virtual training environments to teach behaviors to agents, which conduct actions within the environment to optimize reward functions \cite{ibarz2021train}. Various RL methods have been applied to motion planning applications including actor-critic models \cite{prianto2020path} and deep deterministic policy gradients \cite{li2020motion}. These approaches are most commonly applied to complex transfer motion tasks such as manipulation which are a challenging task for conventional motion planning methods. The definition of the reward function is critical to training the desired behavior and must be carefully designed to ensure training convergence. 
ML-based methods can greatly improve conventional sampling methods as well as serve as superior alternatives when trained appropriately and in the right context, but like all large ML models suffer from challenges of data acquisition. ML models require sufficient training points either through simulation or physical trials which can be incredibly time consuming and resource intensive, especially for DL and RL models. Digital training environments and their capability to transfer to physical implementations remain a bottleneck in model development and must be carefully considered to achieve desirable real-world performance over conventional sampling methods.

Motion planning algorithms which consider constraints found in task motions greatly improve computational efficiency over unconstrained conventional methods \cite{kingston2018sampling} and have been explored both in conventional and ML-based approaches. Motion planner constraints are typically represented as a distance from the surface of a constraint manifold \cite{jaillet2012path}. 
Kingston et al. identified five methodological approaches to handling constraints for sampling-based motion planning: 1) relaxation, where a tolerance is added to the constraint function from which sampling is conducted, 2) projection, where a configuration is projected onto the surface of the implicit manifold and stepped backward into a satisfactory configuration, 3) tangent space, where a tangent space is generated from the constraint function, 4) atlas, where a piecewise linear approximation of the manifold is created using tangent spaces, and 5) reparameterization, where the robot configuration is reparameterized to allow for direct sampling of a satisfactory configuration space. Though not explicitly explored in RAM applications, motion planning methods discussed by Kingston et al. \cite{kingston2018sampling} within these fields offer solutions to task motion planning constraints which can be applied to RAM applications. Furthermore, complex planning environments found in C-RAM systems can be represented as constraint manifolds to improve motion planning capabilities.
Constraint manifolds can also be applied to ML-based motion planners and have been explored by Qureshi et al. \cite{qureshi_neural_2020} who extended their neural manipulation planning method to handle constraints. The neural manipulation planner undergoes a projection-type constraint operation to transform motion plans from the planner to be constraint-satisfactory. Though the application is for manipulation tasks, similar approaches can be applied to industrial tasks such as those in RAM systems. Li et al. \cite{li_constrained_2021} devised an alternative approach for constraint handling motion planning using an actor-critic RL model. The RL model reward function is altered to handle the velocity constraints by incurring a penalty when exceeding the desired velocity when the arm is moving far from the target position. The trained model was able to complete motion planning tasks while maintaining velocity constraints successfully for a simulated space manipulation task. Similarly, this approach can be applied to RAM systems through modified reward functions which ensure motion plans that maximize part quality and minimize the chance of creating a defect.

\subsubsection{Software}

\begin{figure*}
\centerline{\includegraphics[scale=0.5]{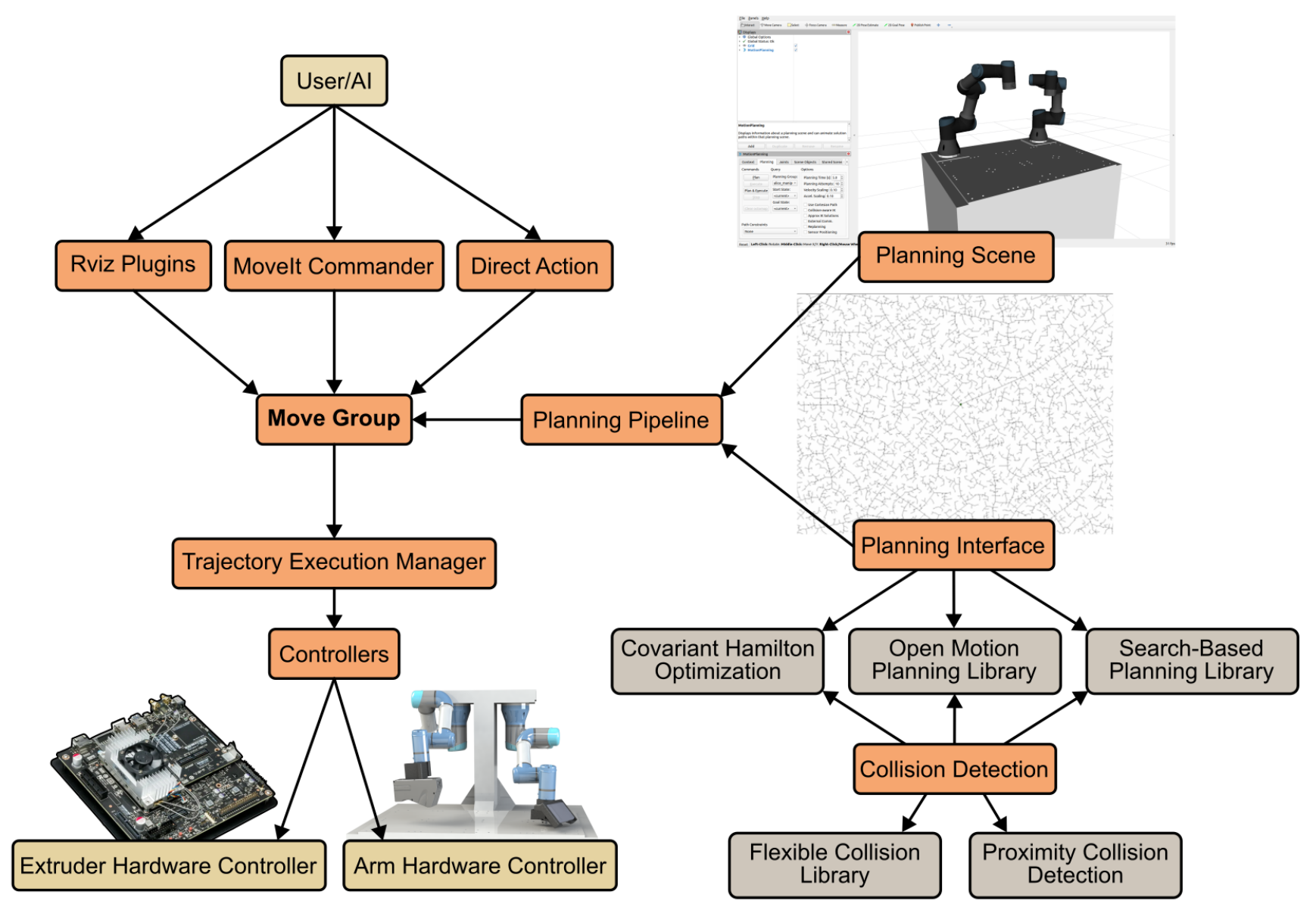}}
\caption{Motion planning frameworks such as MoveIt allow for the integration of various different motion planning methods and libraries (gray) into a ROS2 framework of nodes (orange) to communicate to external hardware (brown) to facilitate RAM (figure derived from MoveIt documentation \cite{moveit_concepts}).}
\label{Fig-ROS-Diagram}
\end{figure*}

Many advancements in software for robotics control have been made in the last two decades for both closed and open-source applications. Software solutions to motion planning must be capable of handling path and trajectory planning as well as collision detection. Closed-source software for control, typically supported by the manufacturer, offers seamless integration with supported machines at the cost of customizability. While most robotic controllers do have the capability to receive commands and conduct motion planning directly, manufacturer-supported software specifically refers to control from an external PC. Open-source software, while less frequently used, allows for configuration but often requires steeper learning curves and is more difficult to integrate. Software used in literature to facilitate control and motion planning of RAM systems is discussed in the following sections in regard to their advantages and drawbacks.

Closed-source software for RAM control is available both from manufacturers and industrial software providers.
As the use of robotic arms within industrial applications has been well established, robot manufacturers have developed their own bespoke software for motion planning and control of their robots. Manufacturer software offers great ease of use at the cost of customization and is most frequently used for fundamental RAM applications which typically use robotic arms from well-established robotics manufacturers \cite{salom2018robotic}. Common robotic manufacturers used for RAM applications include Fanuc, ABB, Yaskawa and Kuka robotic arms \cite{olszewski2020modern}. 
Fanuc offers the ROBOGUIDE software to control Fanuc robotic arms for industrial applications \cite{coletta2022teaching} (i.e., pick and place programming, in addition to the PC Developer's Kit (PCDK) software for more manual control (i.e. direct access to variables, registers, positions, and alarms). While some authors have used PCDK despite its drawbacks (i.e., no motion planning capabilities or collision avoidance) \cite{kajiIntermittentAdaptiveTrajectory2023a, kajiProcessPlanningAdditive2022}, the majority of RAM research using Fanuc arms opt to use custom software besides ROBOGUIDE to facilitate their control \cite{chalvinLayerbylayerGenerationOptimized2020, leeDevelopmentDefectDetection2021}.

\begin{figure*}
\centerline{\includegraphics[scale=0.4]{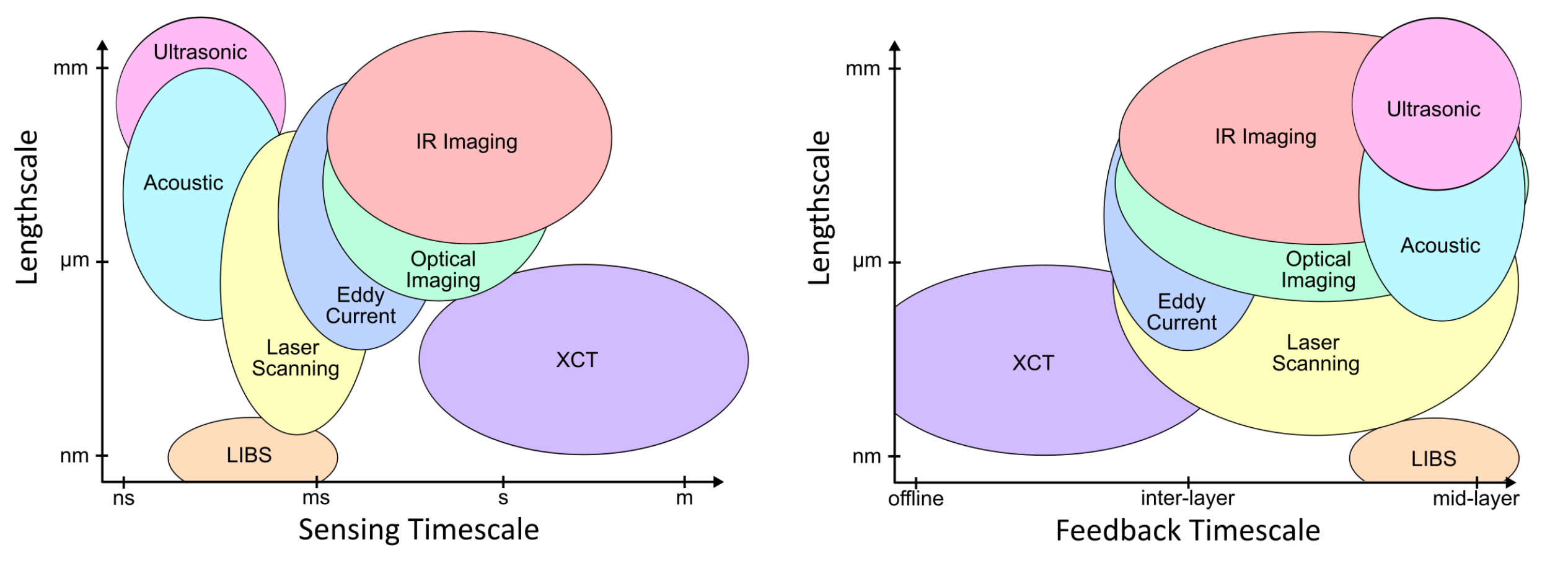}}
\caption{Comparison of sensor fidelity for process monitoring with respect to data sampling frequency and feedback control implementation.}
\label{Fig-ProcessControlLevels}
\end{figure*}

ABB developed the RobotStudio software for control of their robotic arms and supports a variety of features for industrial task planning tasks for both task and transfer motion. 
Closed-source non-manufacturer options include RoboDK which offers CAM capabilities as well as extensions for AM support. 
Open-source software methods afford motion planning solutions with the advantage of being customizable to suit the needs of desired system capabilities while being typically difficult to configure and use compared to closed-source options. 
The most commonly used open-source motion planning solution is ROS or ROS2 (version 2) originally developed by Quigley et al. \cite{quigley2009ros} for prototyping and research applications. It operates as a middleware which supports a large range of packages and libraries to be configured to a user's need and utilizes a publish-subscribe infrastructure agnostic to the robotic system. 
Because of its design philosophy and long development history ROS has seen widespread use in RAM applications.
It is particularly suited for bespoke RAM systems which require customized control such as the integration of sensors.
This framework was later improved upon in ROS2 which was redesigned with production applications in mind to improve major shortcomings in ROS (e.g. not real-time capable) \cite{macenski2022robot}. 
For basic RAM applications, many closed-source and manufacturer-supported software solutions are sufficient for the desired degree of control. However, for bespoke systems that use modified platforms or integrate external systems such as sensors, open-source software gives greater accessibility to pertinent data and control is more suitable. When determining which software to use, understanding this trade-off of initial startup time (which is longer with open-sourced software) and the desired degree of machine control and data accessibility is paramount to realizing the desired system.

\section{Process Control for RAM} \label{sec-ram-defect}


Sensors have been used throughout research in recent years to identify defects throughout all stages of AM processes.
Data collected from sensing data can be further leveraged through the implementation of ML models \cite{razvi2019review}, which are exceptionally well suited to extract features from data collected during AM processes and can be directly linked to defects created mid-process \cite{wang2020machine}. Various types of ML models have been employed for defect detection and quality characterization including hyperdimensional computing \cite{hoang2024hierarchical} and neural network-based architectures \cite{rescsanski2023anomaly}.
Data signals from multiple sensors can also be fused together to further enhance process monitoring capabilities \cite{akhavan2022sensory}. The speed at which inference is conducted affords real time monitoring of defects and control to greatly improve the quality of AM processes.

Feedback control from sensor feedback can occur in three main stages of the AM process: 1) pre-process, where previous sensor feedback is used to inform future processes and adjust parameters before manufacturing, 2) inter-layer, where sensor feedback is used to adjust the manufacturing process in between the creation of layers before the subsequent layer is started, and 3) mid-layer, where sensor feedback is used to adjust the manufacturing process while material is being deposited within a layer. The type of sensors integrated and analysis methods used to interpret feedback dictate the level in which feedback control is feasible and to what degree defects can be rectified within the fabricated component. The advantages and drawbacks of implementations within RAM systems along these process control levels are discussed in the following sections.

\subsection{Pre-Process Calibration}

\begin{figure*}
\centerline{\includegraphics[scale=0.45]{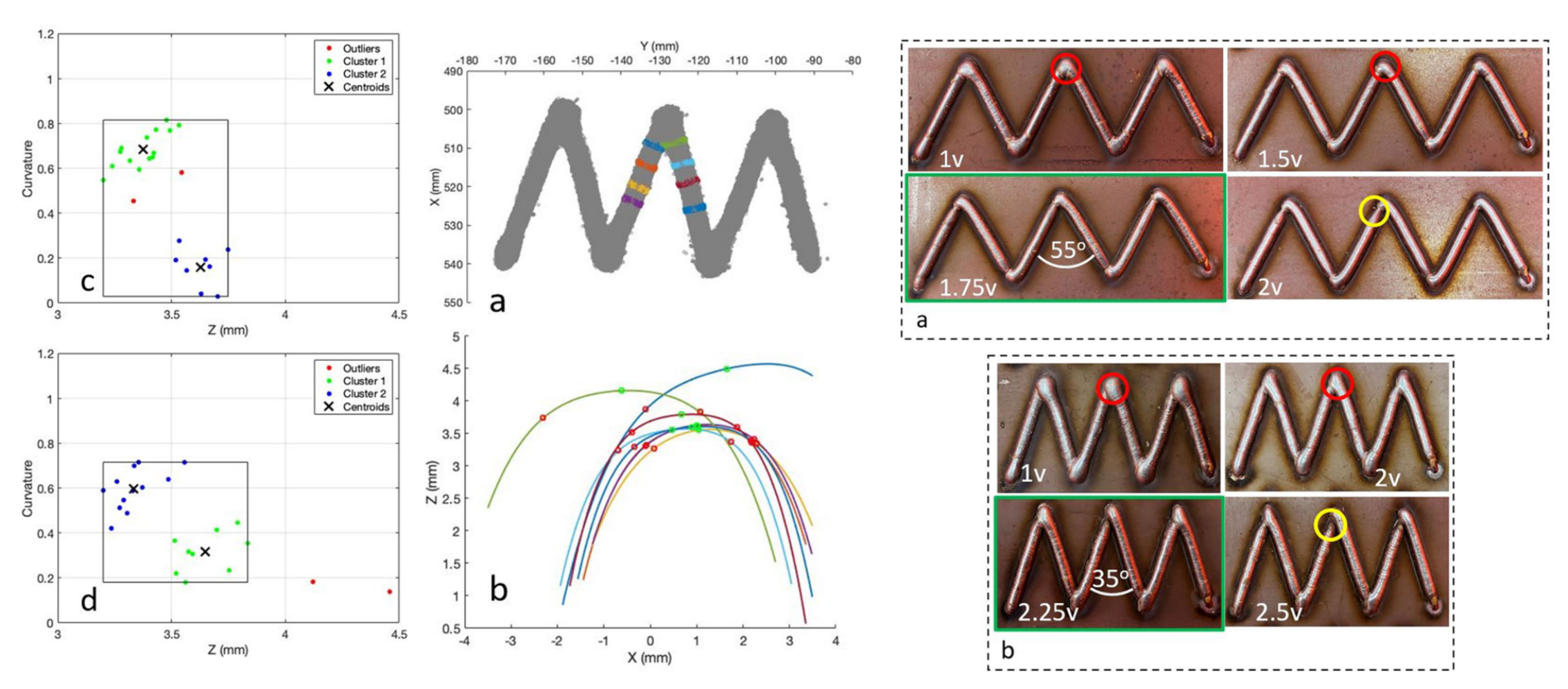}}
\caption{Pre-process calibration methods for RAM systems include toolpath adjustments to prevent defects such as over deposition within sharp corners as proposed by Bhatt et al. \cite{bhattAutomatedProcessPlanning2022}}
\label{fig:preprocess-examples}
\end{figure*}

AM systems including RAM platforms contain many process parameters initialized before the printing process begins which dictate the geometric dimensions and mechanical properties of manufactured components. RAM systems have two categories of process parameters, slicing and manufacturing. 
Slicing parameters impact the generation of the toolpath and include variables such as road width, layer height, infill percentage, and infill pattern. These parameters are typically static.
Manufacturing parameters refer to parameters of the extruder and the AM system and include variables such as feed ratio, feed rate, and extruder temperature. These parameters are set during pre-process but are often modified on the fly. Pre-process operations must determine the variables used during the fabrication process and can be informed in many ways from sensor feedback and data collected to further improve process quality.

Each AM process has specific parameters which modify the function of the AM process and therefore must be optimally selected to achieve the desired process quality and properties of the fabricated part. These process parameters typically control material or energy feed parameters fundamental to the deposition mechanisms of all AM processes.
ME processes such as FFF/FDM which use hot-end nozzles have process parameters including nozzle temperature, bed temperature, and extrusion multiplier (e.g. ratio between extrusion feed rate and rate of TCP motion) \cite{solomon2021review}.
Arc DED processes such as WAAM have process parameters including travel speed, wire feed speed, current and layer height \cite{ty2022influence}.
Laser DED (L-DED) processes have process parameters including laser power and scanning speeds \cite{su2022microstructure}.
As many of these AM processes are well established, previous literature has explored the effect of process parameters through design of experiments as well as ML-guided search methods such as Bayesian optimization \cite{liu2022nonparametric}. In this manner process parameters can be optimally selected before the manufacturing process begins.

Slicing parameters such as layer height, infill percentage, infill pattern, feed rate, and feed ratio are utilized to modify the generated toolpath according to the desired part quality and geometry \cite{garcia2019analysis}. Parameters specific to the AM process utilized are also determined during the slicing process affording further adjustment to the quality of the process and subsequent mechanical properties. According to these selected parameters, the final toolpath generated is then used to create the component. The appropriate selection of these parameters is essential to minimize the prevalence of defects and ensure sufficient mechanical properties for a desired application. As with AM process parameters, many authors have explored the effect of slicing parameters for DED \cite{zhao2020shape} and FFF \cite{sandhu2022influence} including for large-scale systems \cite{rebaioli2019process}. These studies translate to RAM systems directly, though customs systems still typically require calibration and experimental tuning of parameters to obtain good-quality parts.

The pose of the work-piece frame within the build volume of RAM systems can affect repeatability and overall build quality. Determining the optimal pose of the work-piece is paramount to ensuring high quality RAM manufacturing of parts in both single and multi-arm systems. 
Vocetka et al. \cite{vocetkaInfluenceApproachDirection2020} investigated the impact of approach direction on repeatability of an ABB IRB1200 5/0.9 robot with a manufacturer stated repeatability of 0.025mm. A total of 24 target points within the workspace were approached from a sphere of initial positions around the target target position at differing radii, with verification measurements being conducted with a Dantec Dynamics Q-450 with accuracy up to 1µm. Across 24 cases the maximum repeatability improvement from the the best approach direction verse the least advantageous was 0.87mm, displaying the large impact approach direction can have on repeatability.
Ghungrad et al. \cite{ghungrad2023energy} developed an energy-quality map of the workspace such that a part can be positioned to optimally minimize energy usage and positioning errors caused by calibration of D-H parameters. Using the calculated map to determine part placement a maximum energy reduction of 6.5\% and deviation error of 32.7\% was achieved. However, it is unclear how well map generation scales to non-planar and multi-plane printing conditions and non-normal print orientations.
Understanding the relationship between work-piece pose and manufacturing quality is essential to ensure parts are created with satisfactory properties.

\begin{figure*}
\centerline{\includegraphics[scale=0.45]{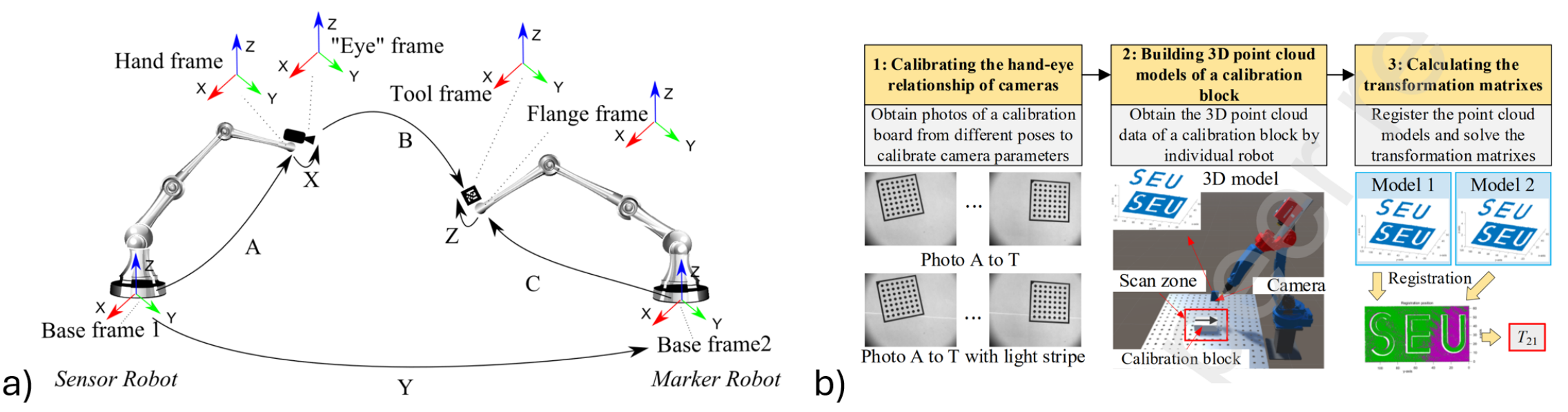}}
\caption{Frame calibration for C-RAM system can be formulated as a) solving $AXB=YCZ$ \cite{wu_simultaneous_2016} and b) can be expanded to use both visual-based calibration and laser scanning for calibration of more than two arms \cite{li_calibration_2024}.}
\label{fig:cram-frame-calibration}
\end{figure*}

Process adjustments are required according to the type of path planning method that is implemented to ensure consistent deposition. 
Bhatt et al. investigated methods to address excess material deposition during sharp corners \cite{bhattAutomatedProcessPlanning2022}. Tight tool path turns result in over deposition during WAAM processes which can be resolved by reducing extrusion multipliers during turns (i.e., increasing end effector motion while maintaining consistent deposition speed). To improve process planning within WAAM systems, a corner detection algorithm is implemented such that the robot feed rate is increased during sharp corners to reduce the amount of material deposited in tight corners. This approach in addition to mid-layer scanning of the surface ensures consistent layer heights and improvements in mechanical properties.
Chen and Horowitz \cite{chenVisionassistedArmMotion2019} investigated ML-based optimization of toolpath planning for an FFF RAM system tasked with fabricating freestanding shapes. Optical images of the printed specimen captured from a static position are processed to extract contour features. A deep deterministic policy gradient (DDPG) algorithm is then trained to maximize the reward defined as the correlation between the printed contour and the desired contour. State actions are defined as sets of points through which the toolhead moves to generate the desired shape.

\subsubsection{C-RAM Frame Calibration}

C-RAM systems operate within a global frame to work on a common part which requires accurate registration of each arm's position. The base frame of each individual robot must be registered within a base frame through a calibration process \cite{wang2015calibration}. 
As multi-arm systems have been used for many years, calibration of arms within a common workspace has been investigated by previous authors using a variety of methods \cite{pradeep2014calibrating}.
The calibration process can be formulated as $AXB = YCZ$ as proposed by Wu et al. \cite{wu_simultaneous_2016} where all variables are homogeneous transforms, specifically the arm 1 base to hand transform $A$, hand to eye (sensor) transform $X$, eye to tool (of arm 2) transform $B$, flange (as opposed to hand) of arm 2 to tool transform $Z$, arm 2 base to flange transform $C$, and arm base 1 to arm base 2 transform $Y$. By knowing transforms of variables $A$, $B$, and $C$ from data acquisition, unknown transforms $X, Y, Z$ can be solved simultaneously \cite{wu_simultaneous_2016}. 

Authors have proposed varying methods to solve the calibration equation in addition to sensing methods to improve the data acquisition process.
Wang et al. proposed a projection-based arm-to-arm calibration method offering an improved maximum error of calibration to previous work using only two calibration points \cite{wang2018plane}. A laser tracker is utilized, and various dual-arm configurations (e.g., floor-wall, wall-ceiling) are evaluated to show generalizability. The method is specifically for dual-arm configurations and would need further decomposition for arbitrary orientation installations which the authors note would require an additional decomposition step. Various sensing configurations have been investigated to achieve similar solutions of the $ABX = YCZ$ equation, 
including optical and calibration target \cite{zhu2018kinematic, zhu2019dual}, optical and ArCo marker target \cite{balanji2022novel}, and binocular vision \cite{ruan2017base}. While these methods successfully solve arm-to-arm calibration they require piece-wise solutions to handle more than two arms in a single system. Additionally, these methods need an additional calibration step to register the workspace (e.g., build platform of a RAM system) with respect to the global coordinates used by the robotic arms. 
To address this, Li et al. \cite{li_calibration_2024} utilized structured light sensors and optical cameras mounted to the end effectors of a 3 arm wire and arc directed energy deposition (WADED) platform to collect point cloud data of a reference part. Eye-hand relationship between the optical camera and the end effector of the arm was characterized for each arm using a calibration board \cite{zhang1999flexible}. Each arm then scans a calibration block using the structured light scanner; these point clouds are then processed using the proposed algorithm to register all point clouds together. From this solution, the corresponding transformation matrices can then be determined to give the final calibration solution for the robotic arms within the system which achieved an orientation accuracy of 0.091°. This approach assumes that there is a high overlap region within the cooperative system which is not necessarily true in large-scale C-RAM configurations. It is unclear how accurate this approach is in large-scale configurations where a minimum amount of overlap is observed, thereby requiring multiple positions from which calibration must be conducted.

\subsection{Inter-Layer Control}

\begin{figure*}
\centerline{\includegraphics[scale=0.5]{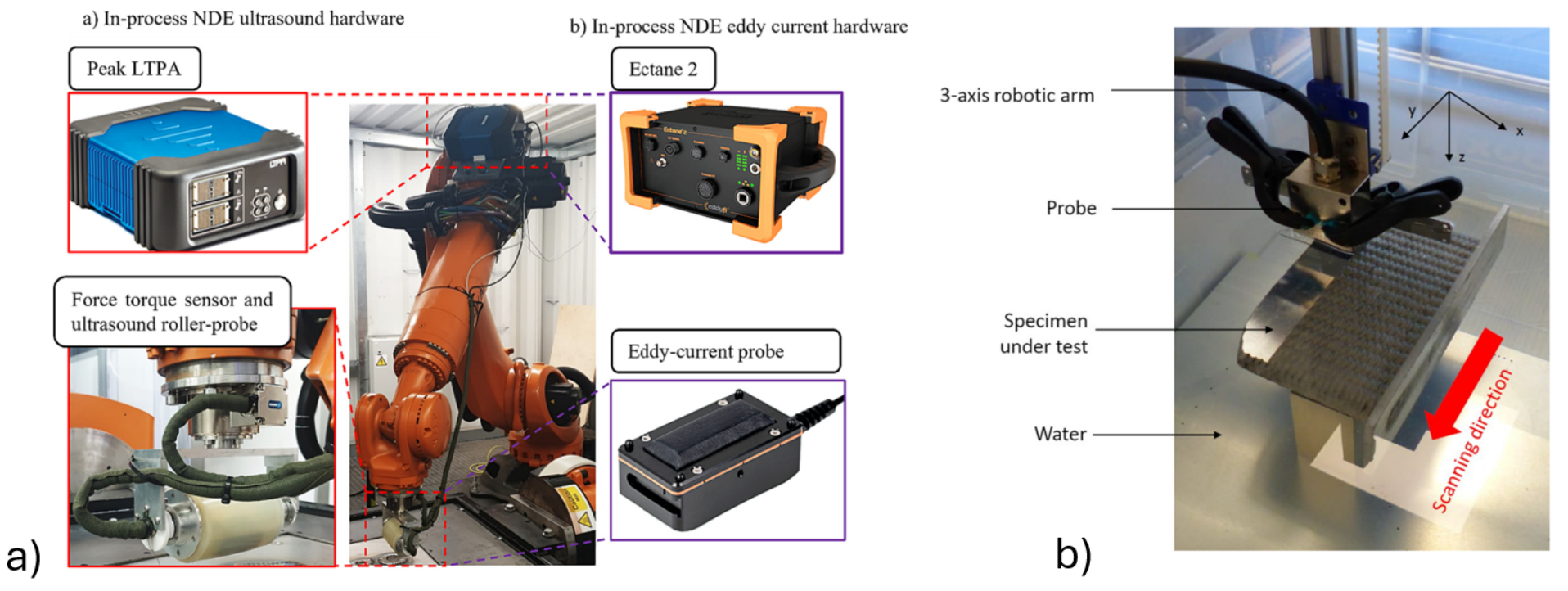}}
\caption{Inter-layer process control typically uses a secondary sensing arm such as a) an ultrasonic roller probe for WAAM \cite{zimermannInprocessNondestructiveEvaluation2023}, b) laser scanners \cite{chabotDefectMonitoringMetallic2020}.}
\label{Fig-Level2}
\end{figure*}

The layer-by-layer nature of AM processes allows for interruption of the manufacturing process for inter-layer inspection and process correction. Feedback can then be leveraged to inform process parameters of subsequent layers minimizing their impact and preventing compounding of defects. For C-RAM systems which use cooperative sensing inter-layer sensing is commonly employed to maximize the scannable region of the part layer. Furthermore, sensors that are too large to mount on the toolhead of fabricating arms can be used in cooperative sensing to offer deeper insight into the quality of the process and defects that are present without sacrificing machine capabilities. Previous literature which employs process control is discussed in addition to the sensing and analysis methods applicable for RAM and C-RAM systems. 

Ultrasonic testing is a class of sensing techniques widely used for DED processes where the porosity of fabricated parts is a pressing challenge.
Chabot et al. investigated implementation of phased array ultrasonic testing (PAUT) in a C-RAM configuration where a secondary 3 DoF arm scans specimens from a WAAM RAM system \cite{chabotDefectMonitoringMetallic2020}. A machined section of an aluminum wall and a propeller blade are evaluated, with example porosity estimates of <0.6mm and <0.6mm being close to digital radiography testing results of 0.69mm and 0.88mm, respectively.   
The authors note that in-situ sensing with an unmachined surface limits the use of this sensing technique to larger pores than the 0.6mm-1mm size range captured in the machined specimen.

Laser scanning offers high-resolution characterization of previously completed layers which can then be used to adjust toolpath plans of subsequent layers.
Magnoni et al. \cite{magnoniRoboticAMSystem2017} implemented an Omron laser triangulation sensor onto the end effector of a 6-DoF FFF IRB-2600 robot arm. After a layer is completed, a post-layer inspection process takes place which follows the original path at an offset height. The measured height is then used to adjust the commanded layer height of the next layer according to the mean hieght measurement. The width and height error from the commanded value was observed to be decreased and correctly controlled leading to overall improved part quality. The method implemented results in geometric deviation from the original CAD design, however, as the layer commands are shifted from their original location with respect to the global origin resulting in a lower overall height.
This drawback was addressed by Rebaioli et al. \cite{rebaioliProcessParametersTuning2019} to include online re-slicing of the original CAD model to retain the original geometry. According to the corrected layer height used in subsequent layers, the remaining region of the CAD model was resliced to ensure that the correct overall height was retained. The drawback of both of these approaches is their coarse method of adjusting the layer height by the nature of using the mean height measurement from the laser scanner. If the layer height is uneven perhaps due to warping or material concentration in differing regions of the part an uneven correction of the layer height would be made. In both examples there are no mention of the variance of the height measurements, making it unclear how well these approaches apply to more commonly found geometries beyond single thin-wall prints.

Structured light scanning can afford similar benefits to line scanners while completing scanning of layers at greatly improved speeds at the cost of measurement precision.  Garmendia et al. \cite{garmendia_-process_2018, garmendiaStructuredLightbasedHeight2019} implemented an HP SLS3 structured light 3D scanner capable of measuring with a maximum precision of 0.05mm in a fixed position above the build plate of a laser metal deposition RAM system. The RAM system consists of a 6-axis ABB 4400 robotic arm equipped with a 2.2kW diode for a 0.6mm spot size working area. Metco 24C martensitic stainless steel powder is deposited via an IK4-TEKNIKER co-axial nozzle onto a C45E carbon steel substrate. Scanning of the previously completed layer is conducted after a set number of layers is completed from which the error of the measured part height to the expected part height can be estimated. If the error is negative an additional layer is repeated to compensate, while if the error is positive a layer from the subsequent planned group is removed. The proposed control strategy is implemented across 0.3mm, 0.7mm, and 1.0mm layer heights with height error being compared with the default no control condition. The proposed methodology was able to reduce layer height error to a maximum of 2mm according to the layer height selected, while the uncontrolled height error displayed compounding height error typical of the AM method implemented. While the proposed methodology showed capabilities of reducing layer height error for L-DED processes, its implementation is limited by the static mounting of the structured light 3D scanner which reduces the effective build volume where the control strategy can be implemented.

Optical imaging has also been implemented to identify defects within the manufacturing process. 
Shen et al. \cite{shenVisualDetectionSurface2020} implemented an end effector-mounted CCD camera in a 6-DoF FFF RAM process to capture surface defects on the exterior of fabricated parts. Images captured in-situ were processed using a dual-kernel method to extract defective regions. The features of the extracted defects are then categorized according to their size and shape into transverse, longitudinal, and localized defects. The sensing implementation does limit the maximum part size due to its interference with the plane of the nozzle, resulting in a negative clearance angle overall. Therefore the methodology is most suitable for cooperative sensing applications rather than single-arm sensing.

Large format sensing devices such as Eddy-Current Testing (ECT) \cite{garcia2011non} can also be employed on secondary robotic arms.
Zimermann et al. implemented a secondary sensing arm consisting of a Kuka KR-90-3100 equipped with an FTN-GAMMA-IP65 SI-130-10 force sensor, ultrasound roller-probe driven by a Peak LTPA controller, and an ED probe controlled via an Eddyfi EC controller for non-destructive testing of a WAAM process \cite{zimermannInprocessNondestructiveEvaluation2023}. WAAM material deposition was achieved through a deposition head consisting of a water-cooled plasma arc welding torch and local shielding device to prevent oxidation mounted on a Kuka robotic arm of the same model. Sensing using the secondary robotic arm was conducted on layers of the WAAM part designated for interlayer cooling, during which the primary arm vacated the build area and a LABVIEW control subroutine was triggered after the temperature of the built part was cooled to operable temperatures for the sensing devices (<150 degrees Celsius for ETC, <350 degrees Celsius for ultrasonic). The proposed methodology was evaluated on a 25mm by 300mm by 25mm straight wall made of Ti-6Al04 V titanium with artificially generated defects created using inserted tungsten tubes and a portable grinding machine to emulate pocket defects. Inspection took less than 1 minute 30 seconds of the 9-minute cooling period for a part 300mm in length with an end effector speed of 0.015m/s for ECT and ultrasonic testing. Each sensing method was implemented on separate layers with ETC testing being conducted on layer 5 and ultrasonic testing on layer 6. Collected data was compared to ground truth x-ray computed tomography (XCT) after the part was completed. The artificial defects were successfully detected by the ECT sensing method while natural porosity defects were not captured. Ultrasonic scanning successfully identified both artificial pocket and natural porosity defects which ranged from 0.1mm to 0.2mm diameter in size. Notably, while ETC sensing requires lower part temperatures than ultrasonic scanning, ETC data can be processed in real time due to its much smaller per-layer data size of 16 megabytes to ultrasonic scanning which uses 1 gigabyte in storage for a single layer. For these reasons, ETC is more suited for low-resolution rapid inter-layer sensing procedures while ultrasonic scanning is more suitable for high-resolution intermediary sensing procedures. Additional details (e.g., how secondary arm tool changes are conducted) and case studies on more complex geometries would clarify the feasibility of the proposed methodology in practical applications of WAAM processes, especially in instances where slicing surfaces are non-planar. 

\subsection{Mid-Layer Control}

\begin{figure*}
\centerline{\includegraphics[scale=0.45]{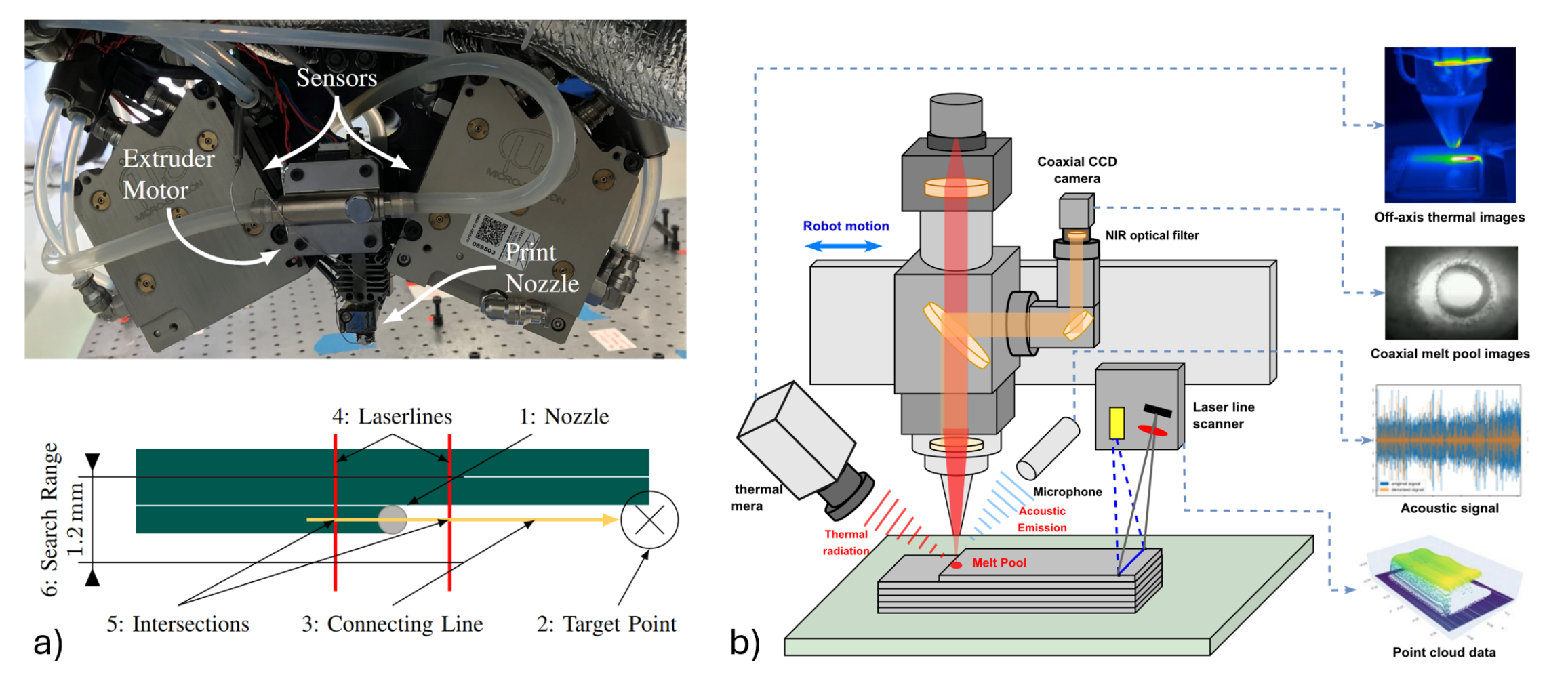}}
\caption{Mid-layer process control for RAM using a) head-mounted laser scanners \cite{mewes_correction_2021} and b) heterogeneous sensing including thermal imaging, coaxial CCD camera and acoustic scanner \cite{chenMultimodalSensorFusion2023}.}
\label{Fig-Level3}
\end{figure*}

Real-time process correction can allow for the mitigation and correction of defects during the manufacturing process. To realize such control equally fast sensing techniques must be employed in addition to analysis techniques that inform closed-loop feedback systems. For AM processes, which create defects in very short time scales (e.g., DED methods including WAAM) implementing real-time process control can realize great gains in mechanical performance and defect reduction. Sensors must also have suitable form factors to be implemented onto the toolhead while minimally reducing the capabilities of the RAM system. Sensors used to facilitate real-time correction during the manufacturing process are discussed at length in addition to the considerations and drawbacks observed in recent literature.

Laser scanners offer high-fidelity sensing at the cost of reduced nozzle clearance for advanced slicing methods.
Mewes et al. integrated dual laser scanners on a 6-DoF RAM system to correct nozzle-bed distance by measuring the height of deposited material and height of substrate in-situ \cite{mewes_correction_2021}.
Kaji et al. utilized laser scanning in-situ, with a Micro-Epsilon scanner mounted offset from a laser DED deposition head on the 6th axis \cite{kaji_intermittent_2023}. In-situ scanning of the material deposited was conducted to allow for defect correction.
Laser scanning is best suited for high-fidelity quantitative characterization of defects which can then be used to infer mechanical properties or correct errors mid process. Data analysis methods must be robust due to the high bandwidth of data,  especially in high-speed fabrication processes or configurations with multiple scanners.

Real-time infrared imaging offers valuable insight into thermal dynamics which can be linked to phenomena such as melt pools. In laser DED processes where a laser is used to melt and deposit material onto a substrate, the associated melt pool can be linked to mechanical properties (e.g., tensile strength) and the generation of defects (e.g., porosity, cracking) according to its size and shape \cite{svetlizky2021directed}. These features can be captured in real time through thermal imaging, and have seen widespread use in DED processes including RAM systems. 
Gibson et al. implemented a melt pool monitoring technique for a laser-wire DED RAM system using an in-axis thermal camera to reduce the impact of wire masking which blocks the view of the melt pool \cite{gibsonMeltPoolMonitoring2019}. This work was then used in a control architecture to control the melt pool size in a DED process \cite{gibsonMeltPoolSize2020}. The proposed method was evaluated across differing controlled variables, specifically measured melt pool size (real-time), average melt pool size (layerwise), and both measured and average melt pool size (layerwise and real-time). Real-time and combined real-time and layerwise control approaches showed the greatest improvement in both bead geometry and thin-wall geometry compared to uncontrolled deposition. This approach is only evaluated for thin walls, however, and it remains unclear how the control scheme works in solid parts with greater variation in thermal gradients and cooling rates. 



Optimal imaging is widely used across AM processes to detect defects ranging from meso to macro scales. 
Many commercially available FFF/FDM gantry systems are equipped with cameras observing build volumes for stringing and delamination defects and have been similarly applied for RAM systems by Badarinath and Prabhu \cite{badarinathIntegrationEvaluationRobotic2021}. The FFF RAM system developed integrated a toolhead mounted camera to live stream build progress for manual monitoring. However, no methods are implemented to automatically detect observable defects such as stringing and delamination. Additionally, there is no indication that the end effector rotates to keep the deposited material in frame meaning not all orientations of deposition are captured.
Conventional image processing techniques such as canny edge detection \cite{canny1986computational} have been widely used to improve feature extraction from optical images. 
Lee et al. \cite{leeDevelopmentDefectDetection2021} implemented an end effector mounted high dynamic range (HDR) camera to capture abnormal metal transfer resulting in hump or valley-type defects in WAAM processes.
Images captured in-situ are filtered using canny edge detection, then input into a convolutional neural network (CNN) architecture to detect valley or hump defects during the manufacturing process. Optimal performance was achieved by a VGG16 CNN model, a commonly used CNN architecture \cite{simonyan2014very}, reaching an overall accuracy of 96.5\%. The humping/valley defect investigated in this study is indicative of improper selection of wire feed speed to torch speed ratio \cite{shah2023review}, however, meaning there is little benefit to online defect monitoring with this approach if proper process parameters are selected.

The use of multiple different sensors in a heterogeneous manner can offer enhanced sensing capabilities able to capture a wide range of dynamics across varying fidelities \cite{rescsanski2022heterogeneous}.
Early examples utilize sensor feedback directly to controllers to improve process quality. 
In one such early investigation, Heralic et al. \cite{heralicIncreasedStabilityLaser2010} implemented coaxial optical imaging to capture melt pool features and a secondary off-axis camera and laser diode to measure layer height of deposition. Measured height is then controlled using a feed-forward compensator, while melt pool width is controlled using a standard closed-loop discrete time PI controller \cite{goodwin2001control}. 
More recent research has sought improved feature extraction and process characterization through ML techniques rather than direct feature extraction methods \cite{chen2021review, shen2022multimodal}. 
Chen et al. implemented coaxial CCD camera, SWIR thermal camera \cite{chenInsituMeltPool2022}, acoustic sensing and inter-layer line scanning \cite{chenRapidSurfaceDefect2021} for a L-DED RAM process \cite{chenMultisensorFusionbasedDigital2023}. Multi-modal data is then processed using a hybrid CNN model to detect keyhole porosity and crack defects \cite{chenMultimodalSensorFusion2023}. While posed as an in-situ sensing technique and showing good results, no discussion related to inference and data processing time is performed to understand how real-time capable the model employed is.

\subsection{Digital Twin}

\begin{figure*}
\centerline{\includegraphics[scale=0.43]{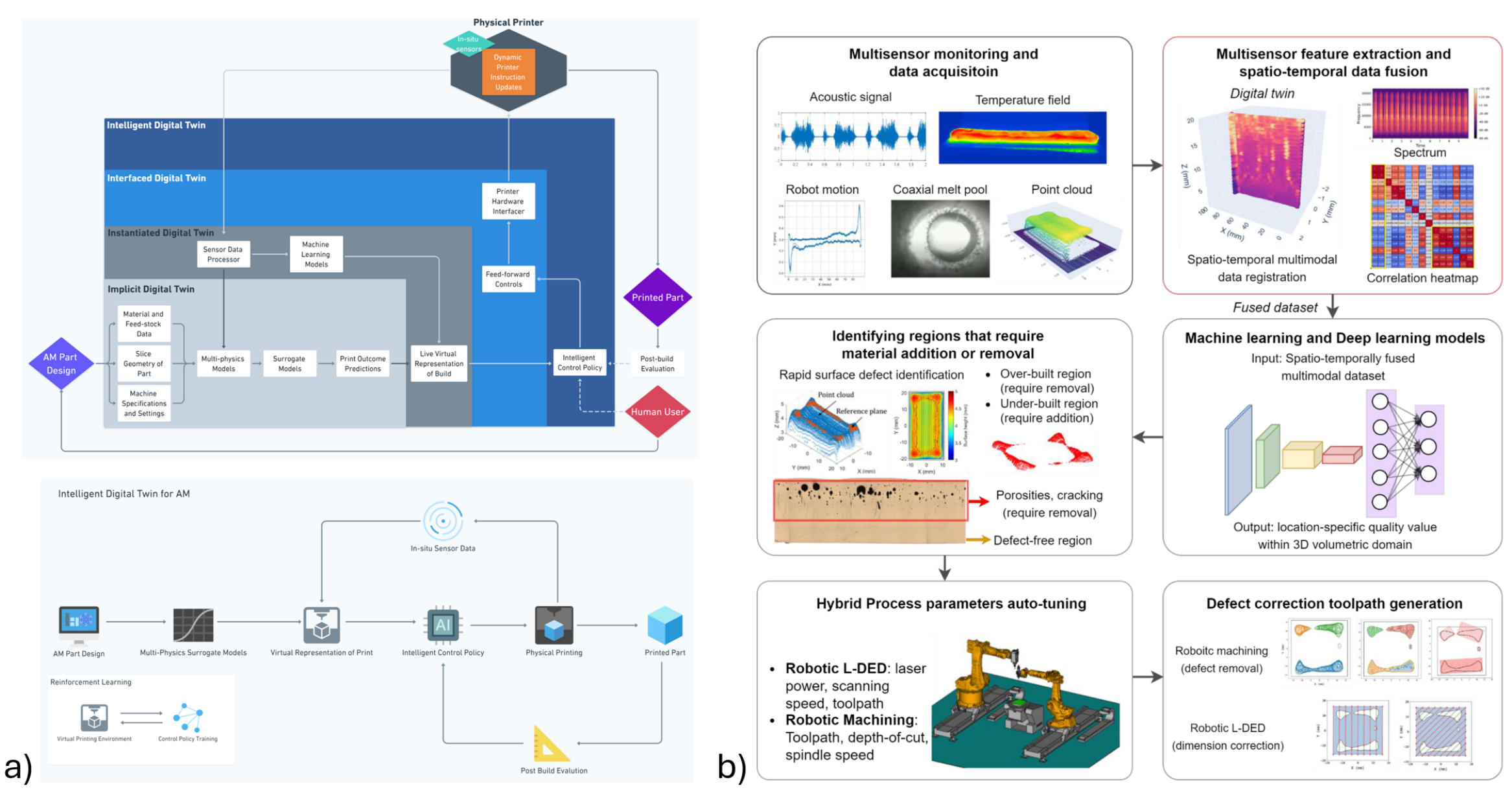}}
\caption{Digital twin structure for RAM processes for  a) WAAM using a hierarchical digital twin structure \cite{phua2022digital} and b) adaptive L-DED AM \cite{chen2023multisensor}.}
\label{Fig-DigitalTwinStructure}
\end{figure*}
Digital twin (DT) of physical systems were first introduced by NASA \cite{tuegel2011reengineering} and subsequently developed for various applications and industries including AM \cite{zhang_digital_2020}. While digital twins for AM is still a developing field, the general approach for digital twins focuses on the use of sensor feedback to inform ML models which simulate physical phenomena using both real-time and historical data \cite{jones2020characterising}. Digital models of the physical system are updated according to sensor feedback in a closed loop fashion to improve approximations and predictive capabilities. Digital models of the physical system are then used to monitor system quality, control processes to improve process quality and predict physical phenomena and their impact on the mechanical properties of the finished component \cite{mu_toward_2023}.
Digital twin applications to metal AM processes have been at the forefront of DT development as a result of its complex microstructures and greater utility (e.g. improved mechanical properties). 

Digital twin systems fall into three subcategories \cite{kritzinger2018digital}, digital model (DM), digital shadow (DS), and digital twin. DMs represent a physical system without any real-time sensor feedback (i.e. it can be constructed based on real-world data). DSs represent the physical system with real-time sensor feedback in a one-way information flow where the digital system is not utilized to control the physical system. DTs use real-time sensor feedback to control the physical system, not just update the digital representation. While many papers are referred to as DT platforms, the degree to which the digital model is updated and integrated into control dictates which subcategory it fits within. 
Furthermore, the resolution and the degree to which the DT characterizes the AM process are useful in distinguishing limitations in approaches. AM process modeling can be subdivided according to the micro, meso, and macro-level physical processes that are modeled. Micro-level physical modeling estimates physical processes such as microstructures, residual stress, and coalescence. Meso-level physical modeling estimates features including cracks, voids, and geometric deviation. Macro-level physical modeling captures part-level characteristics such as Young's modulus, fatigue life, and ultimate tensile strength. Low-level physical modeling can be used to estimate higher-level properties at higher accuracy but typically requires finite element analysis (FEA) and other thermal modeling techniques which are extremely computationally expensive. High-level modeling requires less computation and sensing capabilities at the cost of lower estimation inaccuracies.


Single-arm RAM systems have seen various applications of digital twins to mitigate the impact of defects, especially for metal AM processes such as DED. 
Chen et al. \cite{chen_multisensor_2023} designed a heterogeneous sensing digital twin framework for in-situ monitoring of a hybrid C-RAM system comprised of an L-DED and conventional milling process. The L-DED system is equipped with coaxial CCD imaging, acoustic sensing, and thermal imaging for in-situ monitoring as well as line scanning for inter-layer sensing. The digital twin framework is developed in ROS with data sampling limited to 250Hz to ensure synchronization across axis position measurements and implemented sensors. The authors present a digital framework based on location-specific registration of ML model feedback from sensing data which can then be used to inform parameter adjustment and material correction conducted by the secondary milling robotic arm. However, the framework is only outlined as the authors display preliminary results of localization of quality estimates from sensing feedback. More work must be conducted to fully realize the DT framework proposed to make meaningful use of the quality metrics displayed. 
While there are many examples in the literature of DTs for conventional ME systems \cite{pantelidakis2022digital, guo2021development, kantaros20213d}, to the author's knowledge there are no examples of applications to ME RAM systems. This follows the more general trend of more RAM investigation in DED systems over ME, with processes such as WAAM vastly outnumbering ME. Additionally, DT applications to metal AM processes have been of greater interest \cite{zhangDigitalTwinsAdditive2020}.

\section{Challenges and Opportunities}
While RAM systems have seen great advancements in the last decade, many challenges still persist, which limit the capabilities and application of these systems for both DED and ME configurations. From this review of recent literature, the authors note 5 specific gaps that can be expanded upon in future research: unified slicing software, informed slicing, intelligent C-RAM systems, informed part decomposition, and digital twin software.

\begin{itemize}

\item \textbf{Unified Slicing Software:}
Many authors have investigated advanced slicing methods for many years, but progress remains limited by the inability to replicate and recreate slicing methods. Conventional slicing software follows paradigms that were originally developed for single-extruder 3 DoF platforms, which require further modification to support multi-plane, non-planar, or multi-extruder printing. These extensions depend greatly on the robotic arm used due to the requirement of motion planning solutions and variation in system configuration (e.g., placement of robot arms, location of build platform, extruder configuration, DoF). Many authors have sought to develop bespoke slicing software that further modifies existing slicing software or acts as standalone programs suitable for their experimental setup and are therefore unusable for other systems. 
The continued advancement of RAM systems is dependent on building upon state-of-the-art slicing methods to realize the full potential to improve AM regarding part quality, maximum build size, and fabrication speed. While no standard library or software exists to the author's knowledge, particularly for RAM systems, software such as the $S^3$ slicer by Zhang et al. \cite{zhang2022s3} have been developed by researchers as open-source solutions. These software in addition to conventional gantry software such as slic3r and reprap \cite{ranellucci2013reprap} can act as building blocks to expand comparability between slicing methods to improve AM research and applications of AM methodologies to other fields. A standard software has yet to emerge that alleviates current drawbacks and potentially revolutionizes research and greatly accelerates its progress.
 
\item \textbf{Quality Informed Slicing:}
Advanced slicing methods including non-planar and multi-plane have been investigated in DED and especially ME processes, but few studies have investigated their impact on the mechanical properties of the slicing method selected. While previous research has investigated approaches to optimize metrics including support material reduction \cite{wu2019general} and tensile strength \cite{fangExceptionalMechanicalPerformance2023}, the impact of advanced slicing methods on the generation of defects (e.g., porosity, voids, cracking, delamination, overfill/underfill) has not been studied. Challenges of motion planning associated with RAM further complicate the generation of defects for complex toolpaths which can have challenging path planning solutions compared to gantry-style platforms. Investigation in this area is limited in part to the slicing methods that are developed for specific experimental setups and their inaccessibility, which prevents straightforward comparative analysis. If the connection between slicing parameters of advanced slicing methods is understood, an informed toolpath that minimizes these influences can be obtained. As many quality issues still exist within RAM systems, it is therefore paramount to address these drawbacks and realize the full potential of these platforms.

\item \textbf{Intelligent C-RAM Systems:}
Many fabrication C-RAM systems have been developed to solve large manufacturing limitations of conventional AM systems, but have not investigated the mechanical properties of printed components. Handling the motion planning of multiple arms requires fabrication while also preventing collisions which can alter the ideal toolpath plan and cause varying cooling rates within the part. These factors are all affected by the selection of the slicing and motion planning method and their impact can be quantified through the use of previously developed in-situ monitoring techniques. 
To ensure parts are created with desirable mechanical properties and minimal defect, process monitoring through sensing should be investigated as conducted in previous literature. Specifically, sensing integration can be used to quantitatively assess C-RAM defects including within segment boundaries to inform the toolpath generation and motion planning process which must jointly consider the impact of each arm. The lack of research within this area indicates a massive area of potential improvement for C-RAM systems both for DED and ME processes.

\item \textbf{Informed Part Decomposition:}
C-RAM fabrication platforms necessarily have segment boundaries located within overlapping build regions for large-scale system configurations. These boundaries present an additional point of failure in both DED processes, where thermal distributions result in residual stresses and warping, and in ME processes, where discontinuous deposition leads to anisotropic properties. Therefore, the determination of segment interfaces and the selection of segment locations is critical to maintaining mechanical properties in parts created with C-RAM systems. While few authors have proposed alternative interface structures, no investigation has evaluated the selection of interface shape and location on the mechanical properties of finished components. If C-RAM systems are to be proposed as a replacement to large-scale gantry systems, which do not suffer from segment interface drawbacks, novel algorithms for interface selection and evaluation of their impact should be developed. Further investigation into the selection and impact of the segmentation process in C-RAM systems would allow for the minimization of segment interface impact and greatly improve the mechanical properties of large-scale AM components.

\item \textbf{Digital Twin for RAM Systems:}
As AM sensing methods and data analysis techniques have greatly improved in recent years, the impact and potential of digital twins for AM have similarly grown. RAM systems which must coordinate the motion of multiple arms in harmony with each other stand to gain the greatest benefit of digital twin software among AM system types. Though digital twin software has been investigated in RAM systems, this area of research stands to gain from standardization and open sourcing of software used to evaluate and implement digital twins for RAM systems. Full realization of digital twin software depends on ML models and sensors which must fulfill minimum requirements of end-to-end latency. These key components should be evaluated comparatively to understand the limitations and gaps of current methodologies. As with slicing software, closed-source solutions limit the advancement of research within RAM and prevent the development of novel systems. These points are especially critical in digital twins which span many areas of research and expertise. Further discussion regarding digital twin's role in AM and generalized approaches to implementation would greatly benefit RAM research and manufacturing development as a whole.

\item \textbf{Motion-Planning Informed Slicing:}
For manufacturing systems (e.g., CNC, FFF, DED) with trivial inverse kinematic solutions (e.g., gantry) toolpath planning implicitly solves the motion planning problem due to only a single solution existing. Therefore, conventional toolpath planning software (i.e. slicing software) consider dynamic and geometric limits of the system to ensure an optimal toolpath is generated. This is not true for RAM systems, which typically use g-code generated from slicing software intended for gantries which are then interpreted to machine commands for robotic controllers. This involves simplifications such as assuming end effector orientation according to the normal of the surface or slowing down toolpaths to suit robot motion. Motion planning is then solved separately as an afterthought giving no bearing to the toolpath that is generated. It is known that factors such as stiffness of the robot arm \cite{perez2018study} and approach direction of the TCP \cite{vocetkaInfluenceApproachDirection2020} can affect the repeatability and manufacturing quality of the system. Furthermore, the toolpath remains optimized in regards to jerk of joints and TCP in addition to the maximum operating capabilities of the robotic arm. As a result, toolpaths are unoptimized and slow compared to conventional gantry systems (in regards to feedrate) and do not approach the maximum capabilities of the robotic arms themselves. The conventional slicing procedure for robotic arms should incorporate the motion planning step such that the impact of the toolpath generation considers the impact on motion planning to realize optimized toolpaths for RAM systems. This becomes of further interest for systems with more than 6-DoF which can reconfigure in place while retaining TCP orientation \cite{kong2015type}. Optimal toolpaths should also consider physical limitations of the AM toolhead and issues of singularities \cite{zhangSingularityAwareMotionPlanning2021b}. From these improvements, the full-speed capabilities of robotic arms can be utilized for large-scale part fabrication.

\end{itemize}

\section{Conclusions} \label{conclusion}
RAM systems offer incredible benefits over conventional AM systems including improved build volumes and enhanced capabilities from the cooperative integration of multiple arms into C-RAM systems. Similar to AM processes, the quality of printed parts and the generation of defects remain the primary challenge of RAM processes versus conventional subtractive methods. To mitigate the impact of defects and ensure high-quality AM processes for C-RAM systems various avenues of control and system design must be considered to optimize the desired mechanical properties of finished components. 
These areas fall into two categories: system control and process control.
System control encompasses the generic control of RAM systems, including slicing methods used to generate toolpaths and motion planning methods for high degrees of freedom robotic arms. Process control focuses on specific AM processes, involving sensor-based feedback during pre-process, inter-layer, or mid-layer stages. This feedback can be integrated with digital twin software, enabling predictive control of AM processes.

Many avenues of research exist to bring C-RAM platforms to the next generation of manufacturing to realize their full potential, especially concerning large-scale applications. Challenges regarding the quality and control of multiple robotic arms still pose a primary challenge and have not yet been fully investigated. Furthermore, the use of sensing feedback to inform important decisions such as the interface between cooperative arms has not been leveraged to enable robust printing at large scales. This extends to next-generation software such as digital twins which require sufficient sensing to model the physical environment to allow for predictive control, especially in metal-based processes such as WAAM. This review outlines papers pertinent to these challenges and outlines the gaps and opportunities that exist within these areas to further develop research within AM and advanced manufacturing systems as a whole.

\section{Acknowledgment}
This research is supported by the Graduate Assistance in Areas of National Need (GAANN) Fellowship from the United States Department of Education.

\balance

\bibliographystyle{model1-num-names}

\bibliography{cas-refs}

\end{document}